\documentclass[11pt]{article}

\usepackage[preprint]{acl}
\usepackage{times}
\usepackage{latexsym}
\usepackage[T1]{fontenc}
\usepackage[utf8]{inputenc}
\usepackage{microtype}
\usepackage{inconsolata}

\usepackage{graphicx}
\usepackage{subfigure}
\usepackage{multirow}
\usepackage{pifont}
\usepackage{enumitem}
\usepackage{booktabs}
\usepackage[skins,breakable]{tcolorbox}
\usepackage{adjustbox}
\usepackage{textcomp}
\usepackage{listings}

\definecolor{keywordcolor}{rgb}{0.7, 0.1, 0.1}
\definecolor{tacticcolor}{rgb}{0.0, 0.1, 0.6}
\definecolor{commentcolor}{rgb}{0.4, 0.4, 0.4}
\definecolor{symbolcolor}{rgb}{0.0, 0.1, 0.6}
\definecolor{codebg}{gray}{0.60}

\usepackage{upquote}

\usepackage{amsmath}
\usepackage{amssymb}
\usepackage{mathtools}
\usepackage{amsthm}

\usepackage[capitalize,noabbrev]{cleveref}

\theoremstyle{plain}

\theoremstyle{definition}

\theoremstyle{remark}

\title{SAT-RL: Training Language Models for Logical Reasoning with Symbolic Solvers}

\author{Jialiang Sun \and Kuldeep Meel \\
  University of Toronto \\
  \texttt{sjl@cs.toronto.edu}}

\hypersetup{
  pdftitle={SAT-RL: Training Language Models for Logical Reasoning with Symbolic Solvers},
  pdfauthor={Jialiang Sun and Kuldeep Meel}
}

\begin{document}

\maketitle

\begin{abstract}
Large language models (LLMs) have achieved strong performance in mathematical and logical reasoning, yet they still suffer from the curse of complexity and struggle as problem difficulty increases with the number of constraint interactions and the size of the search space. In contrast, satisfiability solvers such as Z3 provide scalable and verifiable reasoning through constraint solving under formal theories, but lack the flexibility to interpret natural-language problem descriptions. In this work, we propose SAT-RL, a reinforcement-learning (RL) framework that trains LLMs to interact with symbolic solvers through multi-turn tool execution. SAT-RL leverages synthetic data generation with verifiable solutions, supervised warm-up on solver-augmented traces, and solver-grounded reinforcement learning with tool-aware rewards. Experiments show that SAT-RL substantially improves symbolic reasoning with tools on both in-domain and out-of-domain logic tasks. {Our code is available on \href{https://github.com/sunjia72/SAT-RL}{GitHub}.}
\end{abstract}

\section{Introduction}
\label{sec:intro}

Large language models (LLMs) have demonstrated strong capabilities on mathematical and logical reasoning tasks, including the Law School Admission Test (LSAT), the International Mathematical Olympiad (IMO), and broad AGI datasets such as AGI-Eval~\citep{arlsat,deepseekproverv2,agieval}. In particular, reinforcement learning (RL) has played a central role in these advances by enabling long-horizon reasoning, self-verification, and agentic tool calling~\citep{ppo,deepseek-r1,grpo,yu2025dapo}. Despite this progress, recent work on logical reasoning benchmarks~\citep{zebralogic,satbench,enigmata} shows that LLMs suffer from a curse of complexity, in which accuracy drops significantly as the number of constraints and the size of the search space increase. Incorporating symbolic checking further reveals that this accuracy degradation is related to systematic constraint violations and a persistent gap between reasoning rationales and verifiable correctness~\citep{proverqa}.

On the other hand, symbolic solvers such as Boolean satisfiability (SAT) solvers, satisfiability modulo theories (SMT) solvers, and first-order theorem provers (TPs) can scale to large combinatorial spaces and produce reliable logical deductions by finding satisfying assignments or proofs of unsatisfiability under well-defined theories. Commonly used solvers include CaDiCaL~\citep{pysat}, Z3~\citep{z3}, cvc5~\citep{cvc5}, and Vampire~\citep{vampire}, most of which have Python interfaces available through PyPI packages. Inspired by this reliability, a line of research has combined LLMs with symbolic solvers. For example, SAT-LM~\citep{ye2023satlm} and Logic-LM~\citep{pan2023logiclmempoweringlargelanguage} encode natural-language logic tasks into formal constraints and invoke Z3 in a one-shot declarative query, while other systems such as SymPro-LM~\citep{fcorebench} and MCP-Solver~\citep{szeider2024mcp} support agentic multi-turn interactions with solver feedback and self-correction. However, these methods operate only at inference time and rely on solver-specific knowledge implicitly learned during pretraining, making them vulnerable to program errors. Moreover, they offer limited means of aligning solver encodings with the constraints in the original problem statements.

Rather than treating symbolic solvers as fixed inference-time heuristics, a promising research direction is to integrate LLMs with symbolic solvers through training that ensures the reliability of reasoning traces and optimizes interactions between LLMs and symbolic solvers. To close this gap, we introduce SAT-RL, a framework that enables LLMs to plan and iteratively refine symbolic solver calls using reinforcement learning. The framework consists of two major components: (i) a stage for generating synthetic logic data with ground-truth solvers and verifiers to support supervised fine-tuning (SFT) warm-up, and (ii) an RL stage with verifiable rewards and instance-conditioned solver-tool sampling to guide exploration. Overall, SAT-RL aims to produce hybrid reasoning systems that retain natural-language accessibility while achieving scalable, faithful, and verifiable logical reasoning across diverse problem domains.

\section{Related Work}
\label{sec:related_work}

\subsection{Benchmarks for Logical Reasoning}
Benchmarks for logical reasoning span natural-language comprehension, formal validity, and verifiable constraint puzzles. Reading-comprehension datasets such as LogiQA~\citep{LogiQA,LogiQA_2.0} and ReClor~\citep{ReClor} test multiple-choice logical inference over passages, while CLUTRR~\citep{CLUTRR} targets relational induction over narratives. Formal corpora such as FOLIO and P-FOLIO~\citep{FOLIO,P-FOLIO}, RuleTaker~\citep{RuleTaker}, and LogicNLI~\citep{LogicNLI} emphasize first-order validity or entailment with more explicit logical structure and, in the case of P-FOLIO, proof-chain annotations. A complementary line of work uses programmatically generated puzzles with exact solution checking: SAT-Bench~\citep{satbench} randomly samples SAT formulas and creates instances with LLM annotations, ZebraLogic and BoardgameQA~\citep{zebralogic,boardgameqa} provide scalable generators with difficulty control, and Enigmata~\citep{enigmata} broadens coverage across 36 puzzle families. Recent work emphasizes robustness to memorization through automatic generation in AutoLogi~\citep{autologi} and periodic refreshes in NPHardEval~\citep{nphardeval}, and broadens puzzle coverage with long-tail game settings in HardcoreLogic~\citep{hardcorelogic}. Complementary evaluation efforts such as LogicBench and Multi-LogiEval~\citep{logicbench,multilogieval} assemble multi-step logical test suites, while CFBench~\citep{cfbench} evaluates whether models can satisfy explicit constraints in generated outputs.

\subsection{Logical Reasoning with LLMs}
Prompting techniques, such as chain-of-thought (CoT), tree-of-thought (ToT), and self-consistency sampling, improve decomposition and search over candidate responses~\citep{wei2022chain,yao2023tree,wang2022self}. Several studies focus on coupling LLMs with external symbolic solvers to enable verifiable reasoning steps: SAT-LM~\citep{ye2023satlm} issues a one-shot declarative query to Z3~\citep{z3}, while MCP-Solver~\citep{szeider2024mcp} supports iterative constraint-model construction and validation through a structured interface for Z3, PySAT~\citep{pysat}, and MiniZinc~\citep{minizinc}. FCoReBench~\citep{fcorebench} targets first-order combinatorial reasoning with scalable instances.

RL can improve reasoning by directly optimizing model behavior under preference signals or reward functions. Proximal Policy Optimization (PPO) remains a common RLHF backbone due to its stable on-policy updates in large action spaces~\citep{ppo}. Direct Preference Optimization (DPO) avoids an explicit reward model by optimizing a closed-form objective over pairwise preferences~\citep{dpo}. Group Relative Policy Optimization (GRPO) estimates advantages from within-group comparisons over multiple sampled responses, reducing reliance on value-function learning~\citep{grpo}. Recent variants refine policy-gradient baselines or update rules for long-horizon reasoning, including REINFORCE++~\citep{hu2025reinforce++} and OPO~\citep{opo}. Self-play-based methods, such as SPIN and SPPO~\citep{spin,sppo}, generate preference data from model--model interactions to iteratively strengthen policies. DAPO emphasizes stabilizing long-horizon training through improved optimization and regularization~\citep{yu2025dapo}.

\section{Methods}
\label{sec:methods}
\begin{figure*}[t]
  \centering
  \includegraphics[width=0.9\textwidth]{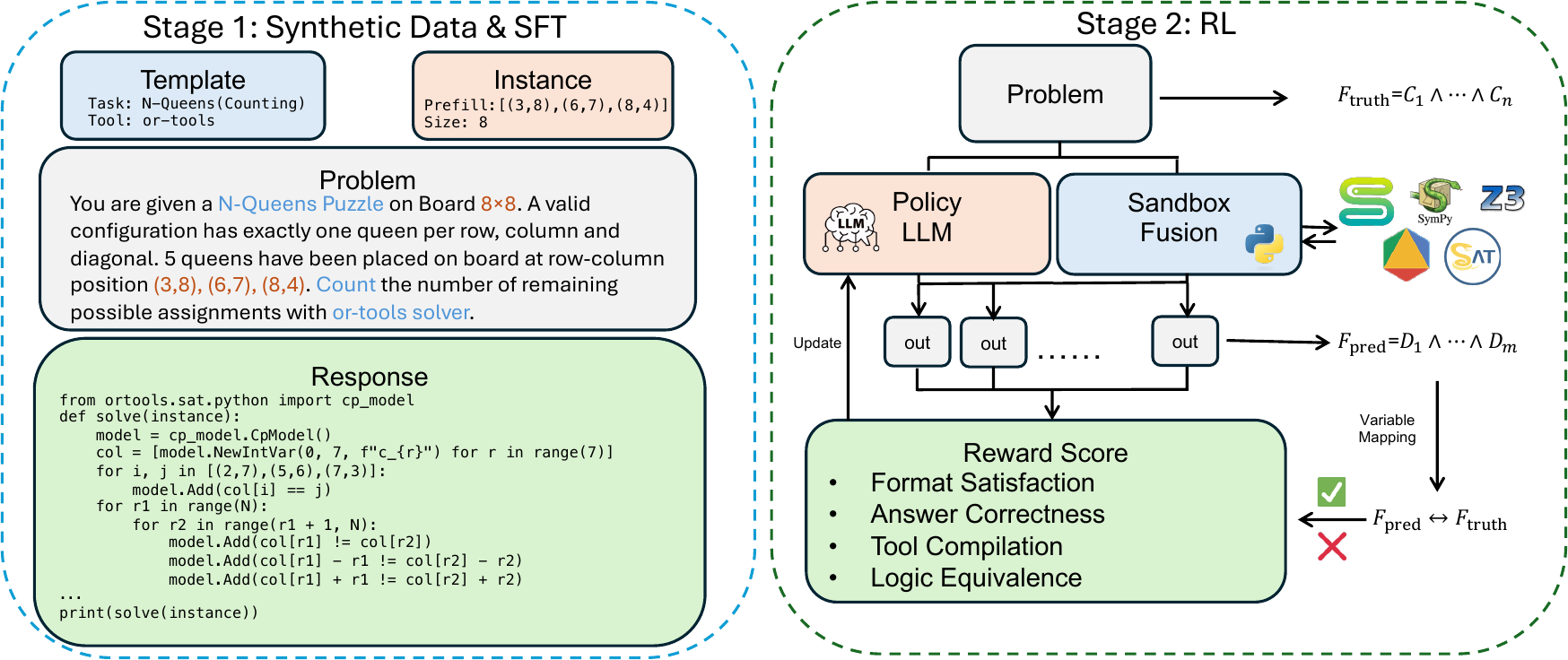}
  \caption{Workflow of the SAT-RL framework. Stage 1 (\textbf{left}) illustrates template-based data generation for the SFT phase: tasks and output programs are instantiated with randomized parameters to produce natural-language problems paired with ground-truth solver programs and executable responses. Stage 2 (\textbf{right}) illustrates the RL training phase and its reward components.}
  \label{fig:workflow_overview}
\end{figure*}

Overall, we integrate long-horizon reinforcement learning with multi-turn tool calls to symbolic solvers. The SAT-RL pipeline comprises three stages: (i) a data generation and preprocessing stage for supervised fine-tuning; (ii) an SFT stage that warms up the base policy model; and (iii) an RL stage that fine-tunes the model using DAPO~\citep{yu2025dapo}, a variant of GRPO~\citep{grpo}, while interleaving code execution via Sandbox Fusion~\citep{sandboxfusion} as the MCP backend. We first introduce the notation and problem formulation and then describe the proposed methods in detail.

\subsection{Preliminaries}
\paragraph{Notation.}
We denote the trainable policy model by $\pi_{\theta}$, where $\theta$ denotes the model parameters. The conditional distribution over outputs is given by
\[
\pi_{\theta}(y \mid x) = \prod_{t=1}^{T} P(y_t \mid y_{<t}, x, \theta)
\]
where $x$ is the input query and $y = (y_1, \ldots, y_T)$ is the output token sequence. 

Throughout the paper, we use the term symbolic solvers to collectively refer to the following PyPI distributions: \texttt{z3-solver}, \texttt{pysmt}, \texttt{python-sat}, \texttt{ortools}, and \texttt{sympy}. In particular, Z3~\citep{z3} and PySMT are SMT interfaces supporting theories such as first-order arithmetic over integers, reals, and bit-vectors; PySAT (python-sat) provides Python bindings to state-of-the-art SAT and MaxSAT solvers; OR-Tools is an optimization library for combinatorial and linear programming; and SymPy is a computer algebra system for symbolic mathematics with limited logical reasoning support. A tool call refers to an executable program generated by the model and executed in a sandboxed environment during rollout.

\paragraph{Problem Formulation.}
\label{sec:formulation}
We study natural-language (NL) logical reasoning tasks that admit a reduction to a satisfiability problem modulo a background theory. Following prior work on solver-grounded reasoning~\citep{ye2023satlm,proverqa}, we formalize each problem instance $P$ as a 4-tuple \(P = (\mathcal{T}, V, \Phi, q)\),
where $\mathcal{T}$ is a first-order background theory, $V = \{v_1, \dots, v_n\}$ is a finite set of free variables, $\Phi$ is a first-order formula over the signature of $\mathcal{T}$ whose free variables are contained in $V$, and $q$ is a query whose free variables are also contained in $V$. The theory $\mathcal{T}$ determines the semantics of the symbols and predicates appearing in $\Phi$ and $q$, including propositional logic, linear integer arithmetic, and linear real arithmetic.

Let $\mathcal{M}_{\mathcal{T}}(\Phi, V)$ denote the set of $\mathcal{T}$-models that satisfy $\Phi$ and interpret exactly the variables in $V$:
\(
\mathcal{M}_{\mathcal{T}}(\Phi, V)=\{\, M \mid M \models_{\mathcal{T}} \Phi \;\text{and}\; \mathrm{dom}(M) = V \,\},
\)
where each model $M$ assigns values to the variables in $V$ that satisfy $\Phi$. We consider the following query types:

\begin{itemize}
    \item \textbf{Decision.} The decision task $\text{Decide}(P)$ asks whether $\mathcal{M}_{\mathcal{T}}(\Phi,V)$ is non-empty. The output is \texttt{True} if and only if at least one model exists.
    \item \textbf{Solving.} Given an optional optimization objective $\mathcal{O}$, the solving task $\text{Solve}(P,\mathcal{O})$ asks for a model $M \in \underset{M' \in \mathcal{M}_{\mathcal{T}}(\Phi,V)}{\operatorname{argmax}}\mathcal{O}(M')$, or any model $M \in \mathcal{M}_{\mathcal{T}}(\Phi,V)$ if no objective is given. Note that when $\Phi$ is a Boolean formula in conjunctive normal form (CNF), a typical MaxSAT problem sets $\mathcal{O}$ to be the number of satisfied clauses, whereas a SAT problem does not specify an objective.
    \item \textbf{Counting.} The counting task $\text{Count}(P)$ asks for the number of satisfying assignments, which equals the size of the model set \(\bigl|\mathcal{M}_{\mathcal{T}}(\Phi,V)\bigr|\).
\end{itemize}
\paragraph{Example.}
Consider the question: In how many distinct ways can a factory package exactly $50$ widgets using only 4-widget and 6-widget boxes?
To model this instance using the 4-tuple $P=(\mathcal{T}, V, \Phi, q)$ defined above, let $\mathcal{T}$ be linear integer arithmetic and $V=\{x,y\}$, where $x$ and $y$ represent the numbers of 4-widget and 6-widget boxes, respectively. Define
$\Phi$ as $(4x + 6y = 50) \wedge (x \ge 0) \wedge (y \ge 0)$.
The corresponding counting query asks for $\text{Count}(P)$. Since the satisfying models are \(\{(2,7),(5,5),(8,3),(11,1)\}\), the answer is $4$. A concrete OR-Tools implementation is provided in Appendix~\ref{app:example_ortools}.

\subsection{Data Generation and Processing}
\label{sec:data_generation}

\begin{table*}[ht!]
  \centering
  \small
  \resizebox{0.9\textwidth}{!}{%
  \begin{tabular}{lcccccc}
    \toprule
    \textbf{Resource} & \textbf{Tasks} & \textbf{Controlled Difficulty}& \textbf{Scalable} & \textbf{NP-hardness} & \textbf{Verifier} & \textbf{Solution Program} \\
    \midrule
    SAT-Bench~\citep{satbench}        & 1  & \ding{51}& \ding{55} & \ding{51} & \ding{51} & \ding{55} \\
    ZebraLogic~\citep{zebralogic}           & 1  & \ding{51}& \ding{51} & \ding{51} & \ding{51} & \ding{55} \\
    FCoReBench~\citep{fcorebench} & 40 & \ding{51}& \ding{51} & \ding{55} & \ding{55} & \ding{55} \\
    Enigmata~\citep{enigmata}       & 36& \ding{51} & \ding{51} & \ding{51} & \ding{51} & \ding{55} \\
    Synthetic-Logic (Ours)                     & 48 & \ding{51}& \ding{51} & \ding{51} & \ding{51} & \ding{51} \\
    \bottomrule
  \end{tabular}%
  }
  \caption{Comparison of logical reasoning benchmarks across different features.}
  \label{tab:benchmarks}
\end{table*}
\begin{table}[ht!]
  \centering
  \resizebox{0.7\columnwidth}{!}{
  \begin{tabular}{l r}
    \hline
    \textbf{Data Type} & \textbf{Count} \\
    \hline
    Total & 48 \\
    Number of Tasks & 16 \\
    Number of Queries & 3 \\
    \hline
    \textbf{Primary Task Category} & \\
    Puzzle Games & 5 \\
    Program Verification & 2 \\
    Integer Programming & 4 \\
    Graph Theory & 5 \\
    \hline
  \end{tabular}
  }
  \caption{Task statistics for {Synthetic-Logic}. Task types are assigned based on the primary problem structure, and each task is counted under exactly one category.}
  \label{tab:task_statistics}
\end{table}
In this section, we describe the data processing procedures for SFT and RL training. Our synthetic dataset, Synthetic-Logic, is used to warm up the policy model during the SFT phase. We leverage SAT-Bench~\citep{satbench} for RL training and Enigmata~\citep{enigmata} for evaluation.

\paragraph{Synthetic Data Generation for SFT.}
We construct the {Synthetic-Logic} dataset for SFT warm-up by systematically selecting task families and generating verifiable puzzle instances under controlled difficulty.
Our dataset covers 16 logical reasoning tasks, including 9 drawn from Karp's list of NP-complete problems~\citep{karp2009reducibility}, such as Subset Sum and the Traveling Salesman Problem (TSP). In addition, we include 7 further NP-hard tasks, such as binarized neural network (BNN) verification and N-Queens puzzles, with full task descriptions provided in Appendix~\ref{app:synthetic_puzzles_description}. Each task is instantiated for each of the three query types---decision, solving, and counting---as formalized in Section~\ref{sec:formulation}. Table~\ref{tab:task_statistics} summarizes the coverage of task categories.

For each combination of task type, query type, and symbolic solver, we predefine a structured template consisting of a natural-language problem description and a ground-truth solver program implementing the formal constraint encoding. Ground-truth answers for decision and counting queries are obtained by executing the corresponding solver program, whereas responses to solving queries are validated using task-specific verifiers that check the correctness of candidate witnesses produced by the model. This design ensures that every instance is paired with an executable and verifiable reference trace grounded in the intended solver semantics.

We randomly sample 10 instances for each fixed combination of task, query type, and solver across different difficulty parameter settings. For decision problems, we target a balanced 50\%/50\% SAT/UNSAT split and discard instances that fail due to solver timeouts during ground-truth execution. This procedure yields 1,399 synthetic SFT $\mathcal{D}_{\text{SFT}}$ instances after removing timeout instances, and they are split into 1,260 training and 139 test instances that are fully verifiable with ground-truth responses and verifiers.

\paragraph{Data Preprocessing for RL.}
We focus on logical reasoning datasets whose difficulty and size can be controlled. For RL training, we use SAT-Bench~\citep{satbench}, which consists of logic puzzles automatically generated from CNF formulas. An LLM verbalizes each clause as a natural-language constraint, and each problem poses a yes-or-no decision question asking whether the constraints are satisfiable. Difficulty is controlled at generation time by the number of clauses and variables, with more clauses yielding harder puzzles. SAT-Bench includes a built-in mapping between each sampled formula variable and the corresponding natural-language entity, enabling SAT-RL to compute the logic-equivalence reward during training.

\paragraph{Data Preprocessing for Evaluation.}
For out-of-distribution evaluation, we test the trained models on Enigmata~\citep{enigmata}, a large collection of synthetic and verifiable logical and constraint-based puzzles, such as Tic-Tac-Toe and the 24 Game. Each instance requires the explicit construction of a satisfying assignment and includes a built-in correctness verifier. The final dataset sizes are reported in Table~\ref{tab:size}.
\begin{table}[ht!]
    \centering
    \setlength{\tabcolsep}{5pt}
    \begin{tabular}{c|c|c|c}
        Dataset / Partition 
        & Train & Test & Total \\
        \hline
        Synthetic-Logic (SFT) & 1,260 & 139 & 1,399 \\

        SAT-Bench (RL)  & 1,575 & 525 & 2,100\\
        Enigmata (Evaluation)  & - & 1,901 & 1,901 \\
    \end{tabular}
    \caption{Sizes of the training and test splits for each dataset.}
    \label{tab:size}
\end{table}

\subsection{Supervised Fine-tuning}
\label{sec:sft}
We initialize the trainable policy model $\pi_{\theta}$ by supervised fine-tuning on the solver-augmented dataset $\mathcal{D}_{\text{SFT}}$ constructed in Section~\ref{sec:data_generation}. Each example $(x,y)\in\mathcal{D}_{\text{SFT}}$ pairs an input prompt $x$ with a target response $y$ that includes a valid solver program embedded within \texttt{<tool\_call>...</tool\_call>}.
We train the policy model using the objective:
\[    \mathcal{J}_{\text{SFT}}(\theta)
\;=\;
\mathbb{E}_{(x,y)\sim \mathcal{D}_{\text{SFT}}}
\Biggl[
\sum_{t=1}^{|y|}
\log \pi_{\theta}\!\bigl(y_t \mid x, y_{<t}\bigr)
\Biggr]\]

This warm start teaches $\pi_{\theta}$ basic solver-tool API usage, planning, and coding, and stabilizes subsequent RL by reducing early-stage invalid tool executions.

\subsection{Reinforcement Learning}
\label{sec:rl}
We fine-tune the SFT-initialized policy model with DAPO~\citep{yu2025dapo} using multi-turn tool execution.
\paragraph{Tool Call Interface.}
\label{sec:tool-interface}
We follow the ReTool paradigm~\citep{feng2025retool} to interleave natural-language reasoning with executable solver calls. The LLM emits reasoning tokens $r_i$ within \texttt{<think>...</think>} blocks and executable code or CLI commands $c_i$ within \texttt{<tool\_call>...</tool\_call>} blocks. A sandbox compiles and executes $c_i$ and returns the resulting stdout and stderr as feedback $f_i$ within a \texttt{<tool\_response>...</tool\_response>} block, which immediately follows the corresponding \texttt{<tool\_call>} block. The rollout thus interleaves reasoning, tool calls, and tool responses as
\begin{equation*}
[r_1,\, c_1,\, f_1,\, \dots,\, r_n,\, c_n,\, f_n,\, a]
\end{equation*}
where $a$ is the final answer.

We implement training using the verl~\citep{verl} pipeline with agentic mode enabled. For multi-turn tool execution, we deploy an asynchronous code server via Sandbox Fusion to support parallel, high-throughput solver execution. We use the prompt in Appendix~\ref{app:prompt} to incentivize the policy model to use symbolic solvers and align variables with the problem context through the built-in variable mapping in SAT-Bench~\citep{satbench}.

\paragraph{Reward Design.}
Overall, we aim to incentivize the policy model to solve diverse logical reasoning problems by learning to query various symbolic solvers using faithful formula encodings. The reward should focus not only on the correctness of the final answer but also on the quality of the encoded formula. Thus, we define the total reward $r$ as the sum of four verifiable components: format $r_{\text{format}}$, correctness $r_{\text{correct}}$, compilability $r_{\text{tool}}$, and quality $r_{\text{equiv}}$. All components are initialized to 0.

For each instance, we uniformly sample one of the five solvers (Z3, PySMT, PySAT, OR-Tools, and SymPy) as the required solver, thereby familiarizing the model with diverse solvers and avoiding collapse to a single solver backend during training~\citep{collapse}. To enable final-answer parsing, we require the final answer $a$ in the response to be enclosed in triple backticks using the \verb|```answer```| format and assign a format reward $r_{\text{format}}=0.1$ when parsing succeeds. For final-answer evaluation, we reward correct answers, encourage an honest \textit{I don't know} response under uncertainty, and penalize incorrect answers to discourage random guessing. We allow an explicit abstention token (rendered as \verb|```idk```|) and define the adjusted correctness reward as follows:
\begin{equation}
r_{\text{correct}}(a)=
\begin{cases}
1 & \text{if $a$ is correct}\\
0 & \text{if $a$ is \texttt{idk}}\\
-1-\varepsilon & \text{if $a$ is incorrect}
\end{cases}
\end{equation}
Because the SAT-Bench training set has only two possible final answers (yes/no), this design makes uniform random guessing strictly suboptimal in expectation. The penalty for an incorrect $a$ naturally generalizes to $\frac{-1}{n-1}-\varepsilon$ for multiple-choice problems with $n$ choices.

We assign a tool-syntax reward $r_{\text{tool}}=0.1$ if the final tool call executes without compilation errors and uses the sampled solver.
Beyond compilability, we also judge the quality of the solver call based on the logical similarity between the ground-truth formula and the proposed formula extracted from the last tool call. This approach is inspired by a body of work on autoformalization~\citep{leaneuclid,liu2025rethinking}.
Suppose the ground-truth and proposed CNF formulas have aligned variables and are represented by 
\begin{align*}
    F_{\text{truth}}=C_1\wedge\cdots\wedge C_n\\
    F_{\text{pred}}=D_1\wedge\cdots\wedge D_m
\end{align*}
We verify the equivalence $F_{\text{truth}}\leftrightarrow F_{\text{pred}}$ using Z3 to check whether 
\[(F_{\text{pred}}\wedge\neg F_{\text{truth}})\vee (\neg F_{\text{pred}}\wedge F_{\text{truth}})\]
is unsatisfiable. We further assign $r_{\text{equiv}}=1$ if the formulas are equivalent. The overall reward score is therefore \[r=r_{\text{format}}+r_{\text{correct}}+r_{\text{tool}}+r_{\text{equiv}}\]

\paragraph{Training Objective.}
We leverage Decoupled Clip and Dynamic Sampling Policy Optimization (DAPO)~\citep{yu2025dapo}, a variant of Group Relative Policy Optimization (GRPO)~\citep{grpo}, as our RL training algorithm. Given an input prompt $x$, DAPO samples a group of $G$ completions $\{o_i\}_{i=1}^G$ from the old policy $\pi_{\theta_{\text{old}}}$, computes the reward $R_i$ for each completion, and estimates the corresponding advantages:
\begin{equation*}
\hat{A}_{i,t}=\frac{R_i-\text{mean}(\{R_j\}_{j=1}^G)}{\text{std}(\{R_j\}_{j=1}^G)}
\end{equation*}
It then optimizes $\theta$ with the following training objective:

\begin{equation*}
\begin{aligned}
&\mathcal{J}_{\text{DAPO}}(\theta)
={}\mathbb{E}
\Biggl[\frac{1}{G} \sum_{i=1}^G \frac{1}{\lvert o_i \rvert}{} \sum_{t=1}^{\lvert o_i \rvert}
\Bigl(
\min\Bigl(
r_{i,t}(\theta)\,\hat{A}_{i,t},
\\
&\text{clip}\bigl(
r_{i,t}(\theta),\,1-\epsilon_{\text{low}},
1+\epsilon_{\text{high}}
\bigr)\,\hat{A}_{i,t}
\Bigr)
\Bigr)
\Biggr]
\end{aligned}
\end{equation*}
where the expectation is taken over $x \sim \mathcal{D}$ and $\{o_i\}_{i=1}^G \sim \pi_{\theta_{\text{old}}}(\cdot \mid x)$. Here, $|o_i|$ denotes the number of tokens in $o_i$, $\epsilon_{\text{low}}$ and $\epsilon_{\text{high}}$ denote the lower and upper clipping thresholds, respectively, and the importance ratio $r_{i,t}$ is defined as
\begin{equation*}
r_{i,t}=\frac{\pi_{\theta}(o_{i,t}|x,o_{i,<t})}{\pi_{\theta_{\text{old}}}(o_{i,t}|x,o_{i,<t})}
\end{equation*}
Complete rollout examples with reward computation can be found in Appendices~\ref{app:solved_satbench} and~\ref{app:solved_enigmata}.

\section{Evaluation}
\label{sec:eval}
In this section, we evaluate SAT-RL across multiple models and datasets to examine its in-domain performance and out-of-domain generalization. In particular, we design our empirical evaluation to answer the following research questions:

\begin{figure*}[!t]
    \centering
    \includegraphics[width=0.49\textwidth]{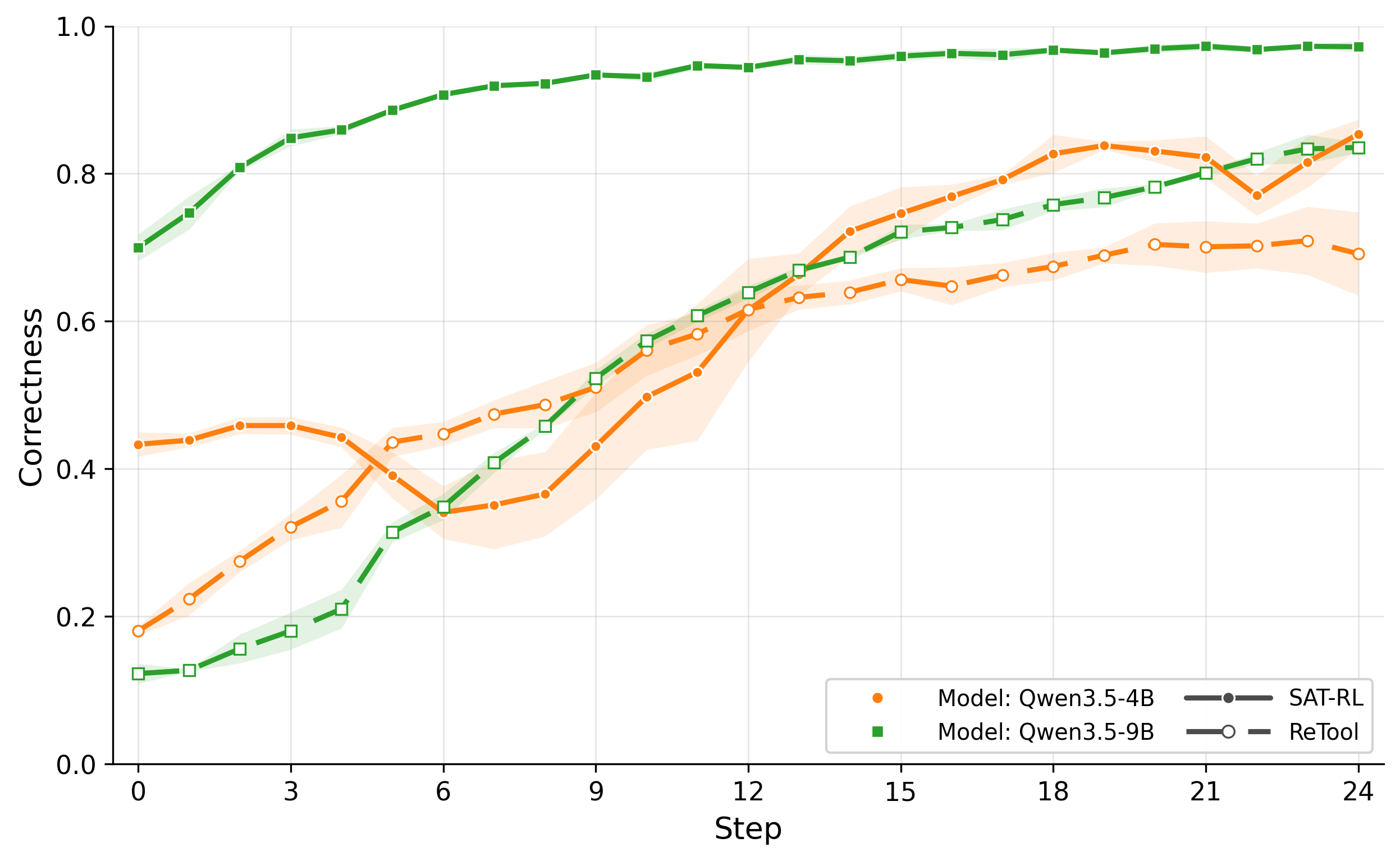}
    \hfill
    \includegraphics[width=0.49\textwidth]{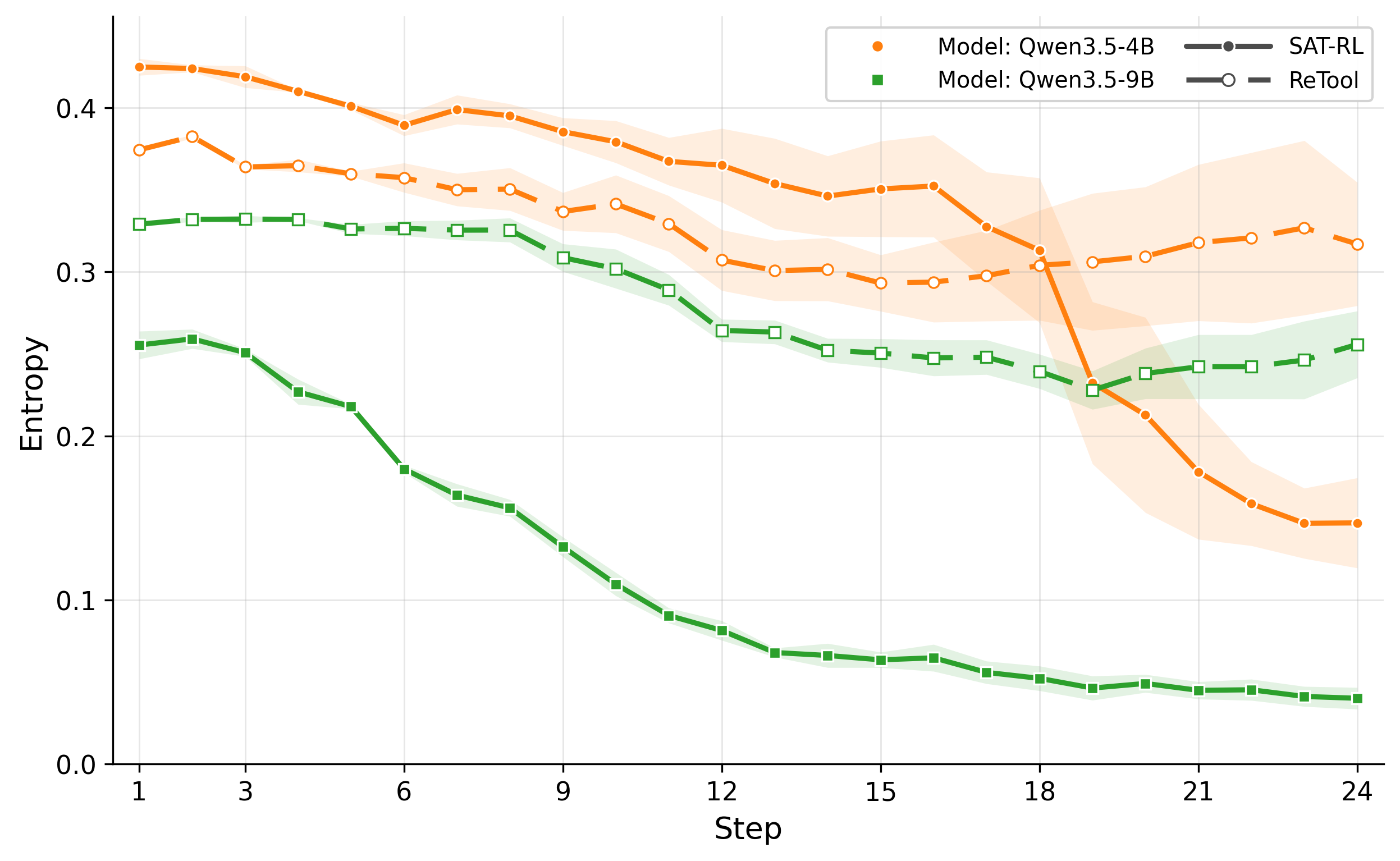}
    \caption{Training trajectories on SAT-Bench for ReTool and SAT-RL (mean $\pm$ population standard deviation across three runs). \textbf{Left}: validation correctness. \textbf{Right}: policy entropy. }
    \label{fig:indomain_accuracies}
\end{figure*}

\begin{enumerate}[label=\textbf{RQ\arabic*.}, leftmargin=*]
\item \textbf{In-Domain Improvements.} Does {SAT-RL} improve accuracy on in-domain logical reasoning benchmarks?

\item \textbf{Out-of-Domain Generalization.} Does training with {SAT-RL} on logic puzzles generalize to out-of-domain logical or mathematical tasks?

\item \textbf{Ablation Studies.} How do key components of {SAT-RL}, including SFT with synthetic data, tool sampling, and logic-equivalence checking, affect test accuracy and training behavior?
\end{enumerate}

\paragraph{Hyperparameter Settings.}
We use verl v0.8.0 as our SFT and RL training framework~\citep{verl}. We use DAPO~\citep{yu2025dapo} with a rollout group size of $G=8$ as the RL training objective. During training, the maximum input and output token limits are set to 4096 and 8192, respectively. The detailed setup is provided in Appendix~\ref{app:setup}.

\paragraph{Model Setup.}
We evaluate cutting-edge proprietary models, including OpenAI GPT-5.6 Luna~\citep{gpt5}, Claude Sonnet 5, and Claude Haiku 4.5~\citep{sonnet5,haiku4.5}, with the reasoning-effort parameter set to ``high.''
We train 4B- and 9B-parameter variants of the Qwen3.5~\citep{qwen3.5} model family.
We compare the SAT-RL regime with the ReTool~\citep{feng2025retool} baseline regime, in which the model can interact with tools without being incentivized to use symbolic solvers, and the reward includes only format, correctness, and successful compilation of tools. We train each model under each regime on the SAT-Bench training split~\citep{satbench} and evaluate in-domain performance on its test split. We use Enigmata~\citep{enigmata} and AIME 2025 to evaluate out-of-domain generalization in logic and mathematics. For each setup, we conduct three training runs $(N=3)$ in parallel with different random seeds to assess training stability. We report test accuracy, with a correct response scoring 1.0 and an incorrect response scoring 0.0.

\subsection{RQ 1: In-Domain Performance}
\label{sec:indomain}

\begin{table}[ht!]
\centering
\setlength{\tabcolsep}{6pt}
\begin{tabular}{@{}lr@{}}
\toprule
\textbf{Model} & \textbf{Accuracy (\%)} \\
\midrule
\multicolumn{2}{@{}l}{\textbf{Base LLMs}} \\
\quad GPT-5.6 Luna & 92.0 \\
\quad Claude Sonnet 5 & 93.0 \\
\quad Claude Haiku 4.5 & 56.0 \\
\quad Qwen3.5-4B & 19.8  \\
\quad Qwen3.5-9B & 14.1  \\

\midrule
\multicolumn{2}{@{}l}{\textbf{ReTool-Trained LLMs}} \\
\quad Qwen3.5-4B & 69.1 $\pm$ 5.7 \\
\quad Qwen3.5-9B & 83.5 $\pm$ 0.7 \\
\midrule
\multicolumn{2}{@{}l}{\textbf{SAT-RL-Trained LLMs}} \\
\quad Qwen3.5-4B (After-SFT) & 43.3 $\pm$ 1.6 \\
\quad Qwen3.5-4B (After-RL) & 85.4 $\pm$ 2.0 \\
\quad Qwen3.5-9B  (After-SFT) & 70.0 $\pm$ 1.8 \\
\quad Qwen3.5-9B (After-RL) & \textbf{97.2 $\pm$ 0.8} \\
\bottomrule
\end{tabular}
\caption{In-domain Avg@8 test accuracy (\%) on the SAT-Bench test set, which contains 525 instances. For trained models, accuracy is reported as the mean $\pm$ standard deviation.}
\label{tab:indomain_merged}
\end{table}

We evaluate in-domain test accuracy on SAT-Bench and observe that SAT-RL enables the Qwen3.5-9B model to outperform cutting-edge proprietary models, as detailed in Table~\ref{tab:indomain_merged} and Figure~\ref{fig:indomain_accuracies}. The base Qwen3.5-9B model performs poorly on the SAT-Bench dataset and has difficulty following the \verb|```answer```| format, with fewer than 20\% of its responses yielding parseable answers. Its test accuracy reaches 70.0\% after SFT and 97.2\% after RL, surpassing Claude Sonnet 5 (93.0\%) and the ReTool-trained Qwen3.5-9B model (83.5\%). A similar pattern appears for Qwen3.5-4B: SAT-RL improves its accuracy from 19.8\% to 85.4\%, outperforming the larger ReTool-trained Qwen3.5-9B model. Overall, SAT-RL yields consistent in-domain accuracy improvements across model sizes.
\subsection{RQ 2: Out-of-Domain Generalization}
\label{sec:outdomain}
\begin{table*}[t]
\centering
\small
\setlength{\tabcolsep}{3pt}
\begin{tabular}{@{}lccccccccccc@{}}
\toprule
\multirow{2}{*}{\textbf{Model}}
& \multirow{2}{*}{\textbf{Overall}}
& \multicolumn{3}{c}{\textbf{Difficulty}}
& \multicolumn{7}{c}{\textbf{Category}} \\
\cmidrule(lr){3-5}\cmidrule(lr){6-12}
& & Easy & Medium & Hard
& Arithmetic & Graph & Grid & Logic & Sequential & Search & Crypto \\
\midrule
\multicolumn{12}{@{}l}{\textbf{Base LLMs}} \\
\quad GPT-5.6 Luna
& 70.9 & \textbf{78.2} & 68.5 & 67.1
& \textbf{100.0} & \textbf{74.5} & 81.4 & 63.9 & \textbf{57.7} & 52.2 & 91.7 \\
\quad Claude Sonnet 5
& \textbf{72.5} & 72.0 & \textbf{72.3} & \textbf{73.2} & \textbf{100.0} & 66.8 & \textbf{86.1} & 75.6 & 52.0 & \textbf{62.2} & 86.7 \\
\quad Claude Haiku 4.5
& 49.7 & 52.0 & 49.7 & 47.8 & 84.2 & 44.1 & 56.6 & 65.6 & 26.1 & 45.0 & 61.7 \\

\quad Qwen3.5-9B
& 33.9 & 47.0 & 33.2 & 24.2 & 70.0 & 37.3 & 42.7 & 45.0 & 18.4 & 23.8 & 15.0 \\
\midrule
\multicolumn{12}{@{}l}{\textbf{Trained LLMs}} \\
\quad Qwen3.5-9B (ReTool)
& 47.9 & 61.6 & 49.7 & 35.2
& 92.5 & 55.9 & 38.6 & \textbf{83.3}
& 22.9 & 41.6 & 77.5 \\
\quad Qwen3.5-9B (SAT-RL)
& 59.9 & 66.6 & 64.8 & 50.1
& \textbf{100.0} & 59.5 & 53.4 & 82.8
& 46.0 & 45.6 & \textbf{100.0} \\
\bottomrule
\end{tabular}
\caption{Out-of-domain Avg@1 accuracies (\%) on the Enigmata test set (1,901 instances), reported overall and across difficulty levels and puzzle categories. Details about the categories are provided in Appendix~\ref{app:enigmata_details}.}
\label{tab:ood_generalization_accuracies}
\end{table*}


We evaluate Qwen3.5-9B on the Enigmata benchmark across difficulty and category partitions after training on SAT-Bench under ReTool and SAT-RL. For each regime, we select the final checkpoint with the highest SAT-Bench accuracy across three runs (Table~\ref{tab:ood_generalization_accuracies}). SAT-RL raises overall accuracy from 33.9\% to 59.9\%, outperforming Claude Haiku 4.5 and the strongest flagship models in categories such as Crypto. Across the easy, medium, and hard subsets, the gains are 19.6\%, 31.6\%, and 25.9\%, respectively. SAT-RL improves all seven categories, with the largest gains on Crypto and Logic, where accuracy reaches 100.0\% and 82.8\%, respectively. These results demonstrate substantial transfer to tasks that emphasize constraint satisfaction and multi-step deductive reasoning.

On AIME 2025 (30 problems), SAT-RL modestly improves Avg@8 test accuracy from 86.7\% (208/240) to 88.3\% (212/240); both the base and SAT-RL-trained Qwen3.5-9B models solve 28/30 problems. Overall, under this configuration, SAT-RL generalizes to out-of-domain logic tasks and provides limited but non-negative gains in mathematical reasoning.

\subsection{RQ 3: Ablation Studies}
\label{sec:ablation}
\begin{figure}[t!]
    \centering
    \includegraphics[width=\columnwidth]{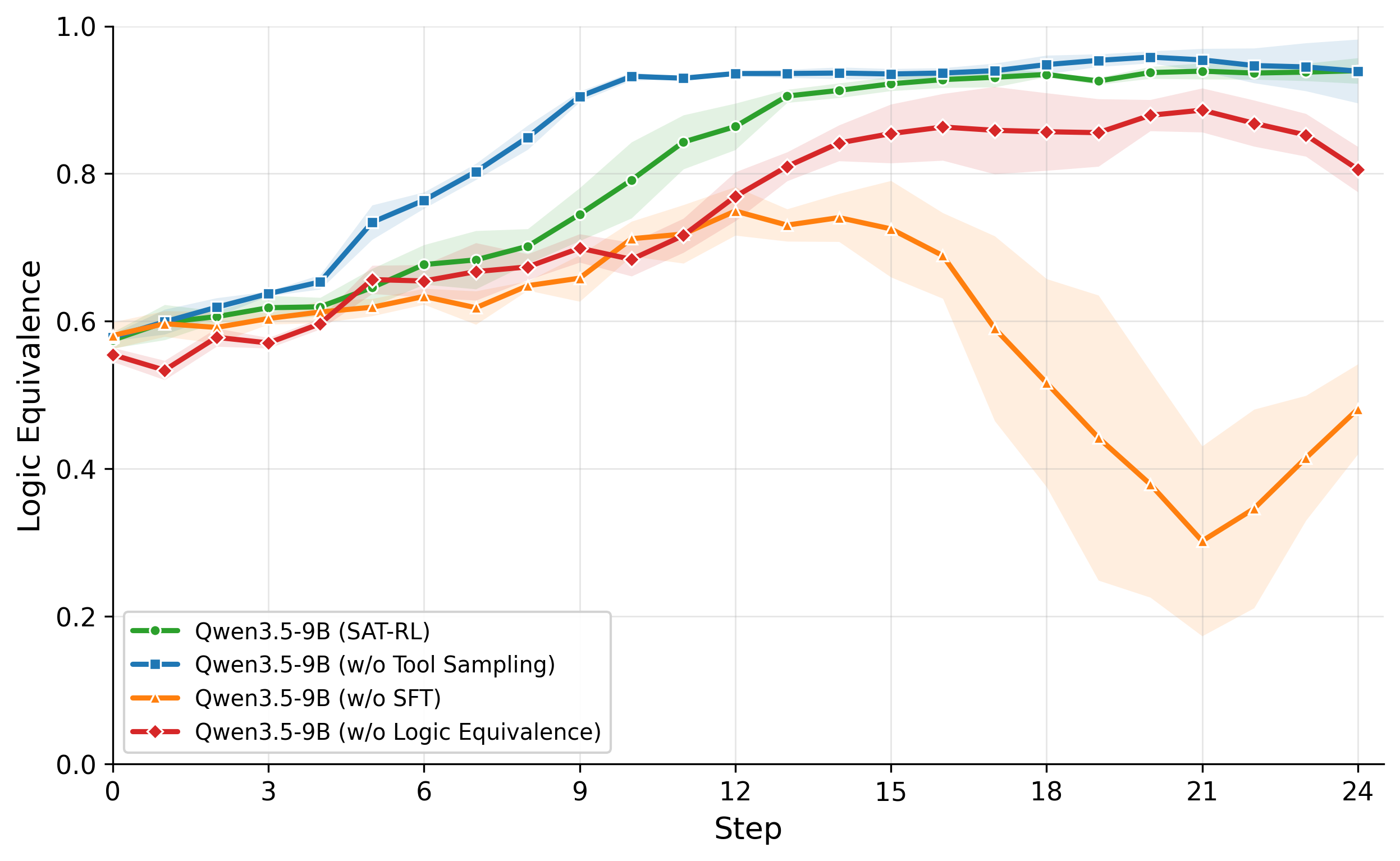}
    \caption{Logic-equivalence trajectories on SAT-Bench. In the ``w/o Logic Equivalence'' ablation, $r_{\text{equiv}}$ is computed but excluded from the training reward.}
    \label{fig:logic_equivalence_ablation}
\end{figure}

\begin{figure}[ht!]
\centering
\begin{minipage}{0.45\columnwidth}
    \centering
    \includegraphics[width=\linewidth]{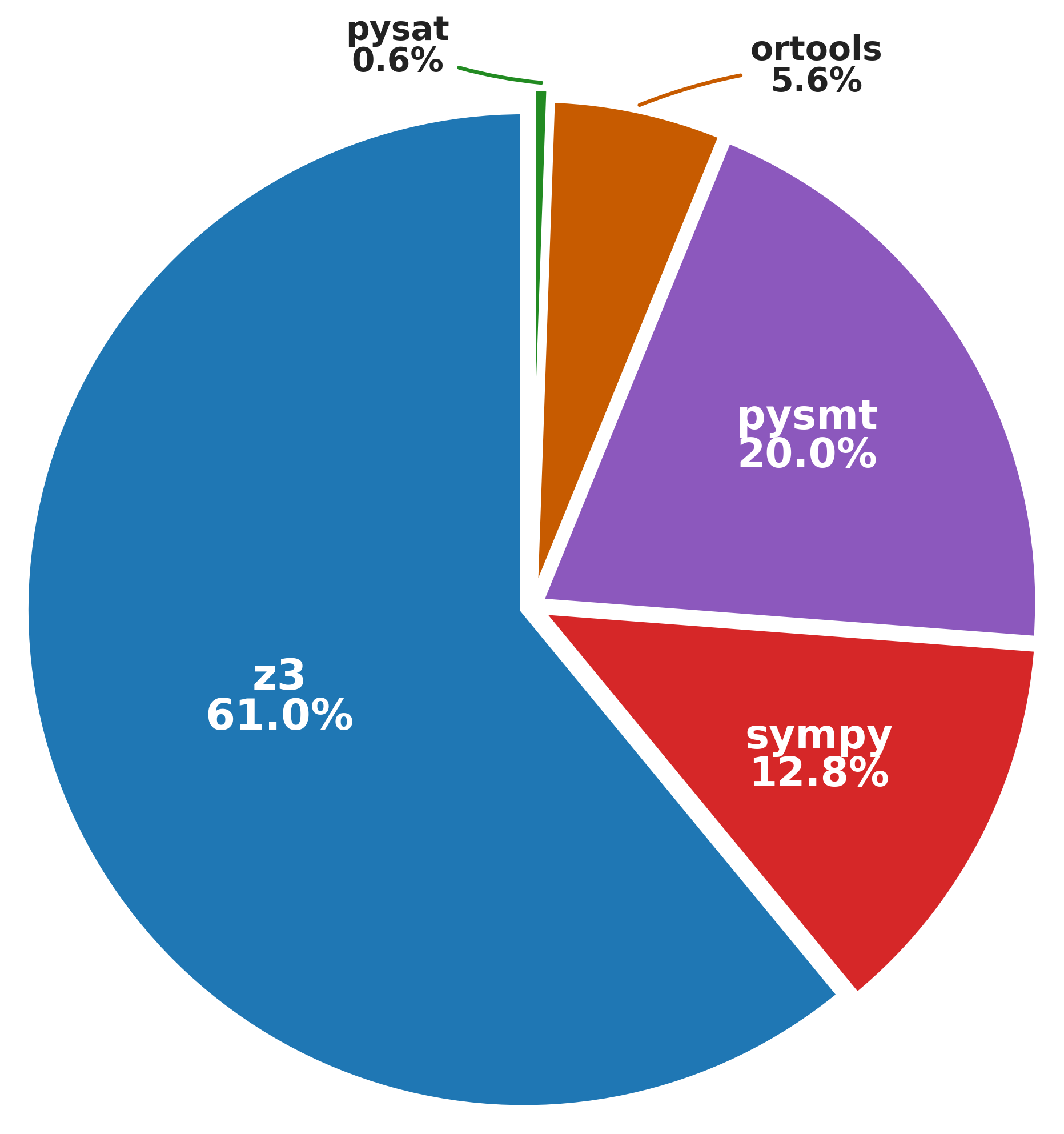}
\end{minipage}
\hfill
\begin{minipage}{0.45\columnwidth}
    \centering
    \includegraphics[width=\linewidth]{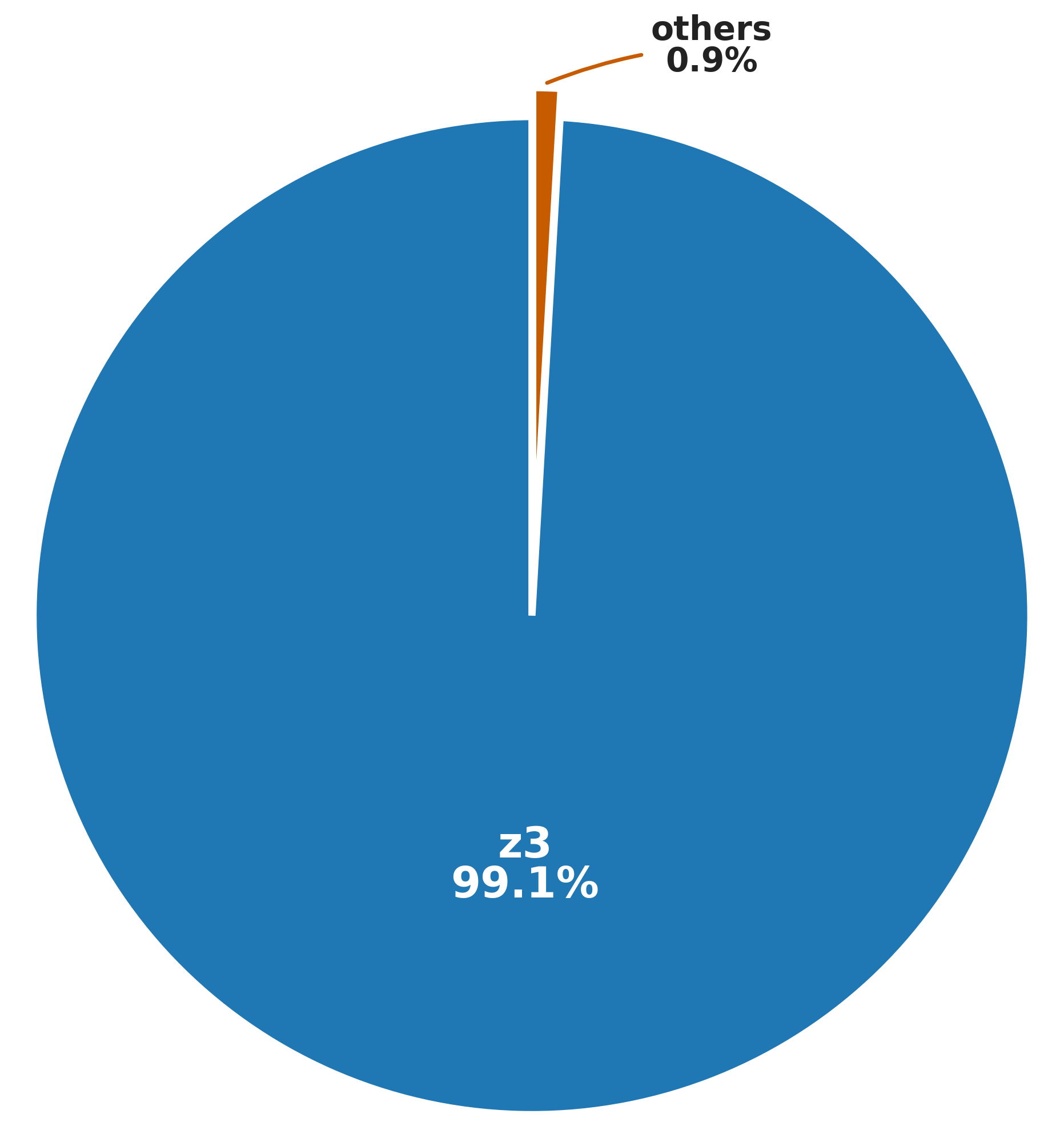}
\end{minipage}
\caption{Distribution of symbolic solver usage with tool sampling enabled (\textbf{left}) and disabled (\textbf{right}).}
\label{fig:tool_diversity_ablation}
\end{figure}

We perform ablation studies to quantify the impact of (i) SFT, (ii) the logic-equivalence reward $r_{\text{equiv}}$, and (iii) tool sampling, as shown in Table~\ref{tab:ablation_satbench}. (i) Removing the SFT stage causes the largest degradation in SAT-Bench accuracy, from 97.2\% to 54.9\%; the latter is comparable to random guessing when responses follow the required parseable answer format. (ii) After removing $r_{\text{equiv}}$ from training, inspection of generated rollouts reveals reward-hacking behavior, including degenerate tool calls such as ``\texttt{from z3 import *\textbackslash nprint(\textquotesingle z3 imported \textquotesingle)}'', which superficially obtain the $r_{\text{tool}}$ reward. This reduces the in-domain accuracy to 91.2\%, with only 80.6\% of solver calls having formula encodings equivalent to the ground-truth formulas. (iii) Ablating tool sampling incurs a slight accuracy drop and causes a severe collapse in solver diversity: over 99\% of tool calls are routed to \texttt{z3}, as illustrated in Figure~\ref{fig:tool_diversity_ablation}. It also reduces Enigmata accuracy to 52.7\%. In contrast, enabling tool sampling yields a substantially more balanced distribution across symbolic solvers. Together, these ablations demonstrate that logic-equivalence checking, SFT, and tool sampling are all necessary for robust end-to-end performance in SAT-RL. Figure~\ref{fig:logic_equivalence_ablation} summarizes the corresponding logic-equivalence trajectories.

\begin{table}[t]
\centering
\begingroup
\setlength{\tabcolsep}{4pt}
\begin{tabular}{@{}lcc@{}}
\toprule
\textbf{Setup} & \textbf{Accuracy} &
{\textbf{Equivalence}} \\
\midrule
\textbf{SAT-RL} & \textbf{97.2} $\pm$ \textbf{0.8} & \textbf{94.0} $\pm$ \textbf{1.7} \\
w/o SFT & 54.9 $\pm$ 8.4 & 48.1 $\pm$ 6.1 \\
w/o TS & 97.0 $\pm$ 2.2 & 93.9 $\pm$ 4.3 \\
w/o $r_{\text{equiv}}$ & 91.2 $\pm$ 2.0 & 80.6 $\pm$ 3.1 \\
\bottomrule
\end{tabular}
\endgroup
\caption{
In-domain final-answer accuracy and logic-equivalence rate of the encoded formulas (\%) on SAT-Bench for Qwen3.5-9B under ablations. ``TS'' stands for Tool Sampling.
}
\label{tab:ablation_satbench}
\end{table}

\section{Conclusion}
\label{sec:conclusion}
We present SAT-RL, an RL framework that integrates synthetic data and symbolic solver interaction into LLM training, enabling accurate logical reasoning and faithful solver use. Experiments show that {SAT-RL} improves in-domain logical reasoning and generalizes to unseen out-of-domain benchmarks, outperforming base LLMs and ReTool baselines.

\clearpage
\section*{Limitations}

Our current study focuses on logical reasoning and a fixed set of widely used symbolic solvers, which limits the scope of this work. Broader program synthesis and open-ended natural-language reasoning remain to be explored. Future work will extend the framework to richer tool-using agent systems and more general coding and mathematical reasoning domains.

\bibliography{references}

\newpage
\appendix

\section{Components of the Enigmata Test Set}
We report statistics for the Enigmata test set (Section~\ref{sec:outdomain}), grouped by difficulty, puzzle category, and task, in Table~\ref{tab:enigmata_instances_by_group}.
\label{app:enigmata_details}
\begin{table}[ht!]
\small
\centering
\begin{tabular}{l r}
\toprule
\textbf{Data Type} & \textbf{Number of Instances} \\
\midrule

\textbf{All instances} & \textbf{1901} \\
\midrule

\multicolumn{2}{l}{\textbf{Difficulty}} \\
\quad Easy   & 560 \\
\quad Medium & 642 \\
\quad Hard   & 699 \\
\midrule

\multicolumn{2}{l}{\textbf{Puzzle category}} \\

\textbf{Arithmetic} & 120 \\
\quad \texttt{countdown} & 60 \\
\quad \texttt{game24} & 60 \\

\textbf{Crypto} & 120 \\
\quad \texttt{crypto\_KKA} & 60 \\
\quad \texttt{crypto\_KPA} & 60 \\

\textbf{Graph} & 220 \\
\quad \texttt{hamiltonian\_cycle} & 60 \\
\quad \texttt{hamiltonian\_path} & 60 \\
\quad \texttt{maze} & 60 \\
\quad \texttt{natural\_language\_navigation} & 40 \\

\textbf{Grid} & 539 \\
\quad \texttt{binario} & 60 \\
\quad \texttt{campsite} & 60 \\
\quad \texttt{magic\_square} & 60 \\
\quad \texttt{skyscraper} & 60 \\
\quad \texttt{star\_battle} & 60 \\
\quad \texttt{sudoku} & 60 \\
\quad \texttt{sudoku2} & 60 \\
\quad \texttt{sum\_skyscraper} & 60 \\
\quad \texttt{full\_crosswords} & 40 \\
\quad \texttt{symbolic\_hard} & 19 \\

\textbf{Logic} & 180 \\
\quad \texttt{FOLIO} & 60 \\
\quad \texttt{knights\_and\_knaves} & 60 \\
\quad \texttt{zebra\_logic} & 60 \\

\textbf{Search} & 320 \\
\quad \texttt{kakurasu} & 60 \\
\quad \texttt{minesweeper} & 60 \\
\quad \texttt{slant} & 60 \\
\quad \texttt{hitori} & 40 \\
\quad \texttt{light\_up} & 40 \\
\quad \texttt{tic\_tac\_toe} & 40 \\
\quad \texttt{checkmate\_in\_one} & 20 \\

\textbf{Sequential} & 402 \\
\quad \texttt{car\_painting} & 60 \\
\quad \texttt{eight\_puzzle} & 60 \\
\quad \texttt{fifteen\_puzzle} & 60 \\
\quad \texttt{nine\_puzzle} & 60 \\
\quad \texttt{sixteen\_puzzle} & 60 \\
\quad \texttt{stack\_permutation} & 60 \\
\quad \texttt{twiddle} & 22 \\
\quad \texttt{big\_bench\_symbolic} & 20 \\

\bottomrule
\end{tabular}
\caption{Number of instances in the Enigmata test set, grouped by difficulty, puzzle category, and puzzle name.}
\label{tab:enigmata_instances_by_group}
\end{table}

\section{Hyperparameter Settings}
\label{app:setup}
The SFT and RL training configurations are shown in Table~\ref{tab:rl_setup}. Training Qwen3.5-9B on SAT-Bench require around 24 hours. 

\begin{table}[ht!]
\centering
\small
\renewcommand{\arraystretch}{1.2}
\begin{tabular}{|p{0.42\columnwidth}|p{0.42\columnwidth}|}
\hline
\multicolumn{2}{|l|}{\textbf{\textit{Software Environment}}} \\
\hline
Python & 3.12.13 \\
{CUDA} Toolkit & 12.9 \\
PyTorch & {2.11.0} \\
{verl} & 0.8.0 \\
{vLLM} & 0.19.1 \\
{transformers} & 5.5.4 \\

\hline

\multicolumn{2}{|l|}{\textbf{\textit{Model and Context}}} \\
\hline
Base Model & Qwen3.5 Model Family ({bfloat16})\\
Max Prompt Length & 4096 \\
Max Response Length & 8192 \\
\hline
\multicolumn{2}{|l|}{\textbf{\textit{SFT Hyperparameters}}} \\
\hline
Train Batch Size & 64 \\
Micro Batch Size & 4 \\
Learning Rate & $1\times10^{-5}$ \\
Total Epochs & 1 \\
\hline
\multicolumn{2}{|l|}{\textbf{\textit{RL Hyperparameters}}} \\
\hline
Train Batch Size & 128 \\
Micro Batch Size (Log Probabilities) & 4 \\
Micro Batch Size (PPO per GPU) & 4 \\
Learning Rate & $1\times10^{-6}$ \\
KL Loss & 0.0 (Disabled) \\
Clip Ratio $\epsilon_{\text{low}}$ & 0.2 \\
Clip Ratio $\epsilon_{\text{high}}$ & 0.28 \\
Clip Ratio $c$ & 10.0 \\
\hline

\multicolumn{2}{|l|}{\textbf{\textit{Rollout Settings}}} \\
\hline
DAPO Group Size ({rollout.n}) & 8 \\
Rollout Backend & vLLM \\
Temperature & 1.0 \\
GPU Memory Utilization & 0.6 \\
\hline

\multicolumn{2}{|l|}{\textbf{\textit{Training Schedule}}} \\
\hline
Total Epochs & 2 \\
Test Frequency (steps) & 1 \\
\hline

\multicolumn{2}{|l|}{\textbf{\textit{Distributed Setup}}} \\
\hline
Actor Strategy & FSDP2 \\
Number of Nodes & 2 \\
GPUs per Node & 4 $\times$ L40S (48 GB) \\
CPUs per Node & 64 Intel Xeon Gold 6448Y \\
\hline
\end{tabular}

\caption{Training configuration.}
\label{tab:rl_setup}
\end{table}

\subsection{Description of Synthetic Puzzles}
\label{app:synthetic_puzzles_description}
\paragraph{Clique (K-Clique).}
Given an undirected graph, the task is to select a set of vertices that forms a clique of a target size. 
Hyperparameters: number of vertices $N$, clique size $K$, and edge list $\texttt{edges}\subseteq \{(u,v):0\le u<v<N\}$.

\paragraph{Feedback Node Set (Directed Feedback Vertex Set).}
Given a directed graph, the task is to remove a small set of vertices so that the remaining graph becomes acyclic.
Hyperparameters: number of vertices $N$, removal budget $K$, and directed edge list $\texttt{edges}\subseteq V\times V$.

\paragraph{Graph $K$-Coloring.}
Given an undirected graph, the task is to assign one of $K$ colors to each vertex such that adjacent vertices have different colors.
Hyperparameters: number of vertices $N$, number of colors $K$, and edge list $\texttt{edges}$.

\paragraph{Max Cut (Threshold Formulation).}
Given an undirected graph, the task is to find a binary partition of vertices whose cut size meets a target threshold.
Hyperparameters: number of vertices $N$, threshold $T$ (required number of crossing edges), and edge list $\texttt{edges}$.

\paragraph{Partition.}
Given positive integers, the task is to split them into two groups with equal sums (equivalently, find a subset summing to half the total).
Hyperparameters: number of items $N$ and the integer array $\texttt{arr}$ (with total sum determining the implied target).

\paragraph{Subset Sum.}
Given positive integers and a target, the task is to choose a subset whose sum equals the target exactly.
Hyperparameters: number of items $N$, target sum $T$, and integer array $\texttt{arr}$.

\paragraph{Traveling Salesman Problem (Threshold Variant).}
Given a complete weighted graph, the task is to find a Hamiltonian tour whose total cost satisfies a threshold constraint.
Hyperparameters: number of nodes $N$, tour-cost threshold $T$, and weight matrix $\texttt{W}\in\mathbb{Z}^{N\times N}$.

\paragraph{Set Covering.}
Given a universe and a family of subsets, the task is to select a small number of sets whose union covers the entire universe.
Hyperparameters: universe size $U$, number of sets $M$, budget parameter $K$, and set family $\texttt{sets}=[S_0,\dots,S_{M-1}]$ with $S_i\subseteq\{0,\dots,U-1\}$.

\paragraph{Set Packing.}
Given a universe and a family of subsets, the task is to select a prescribed number of pairwise disjoint sets.
Hyperparameters: universe size $U$, number of sets $M$, packing size $K$, and set family $\texttt{sets}=[S_0,\dots,S_{M-1}]$.

\paragraph{NFA Path Problems.}
Given an NFA (as an explicit transition relation), the task is to find an accepted path of a fixed length from a start state to an accepting state.
Hyperparameters: number of states $S$, alphabet size $A$, path length $L$, start state $\texttt{start}$, accepting set $\texttt{accepts}$, and transition list $\texttt{transitions}=[(s,a,t)]$.

\paragraph{Binary Neural Network (BNN) Robustness.}
Given a threshold-gate BNN and a seed input, the task is to find an input within a Hamming ball that flips the network output.
Hyperparameters: input dimension $\texttt{n\_in}$, network structure $\texttt{layers}$ (neurons with threshold $k$ and signed literals), and property dict $\texttt{prop}$ containing $\texttt{seed}$, radius $\varepsilon=\texttt{eps}$, and $\texttt{y\_seed}$.

\paragraph{Partial N-Queens.}
Given a partially filled $N\times N$ chessboard, the task is to place one queen in each row while respecting the pre-placed queens and nonattacking constraints.
Hyperparameters: board size $N$ and fixed queen positions $\texttt{pre}=[(row,col)]$.

\paragraph{Generalized Sudoku.}
Given an $N\times N$ Sudoku grid (with $N=n^2$), the task is to fill in the blanks while satisfying row, column, and sub-block uniqueness constraints.
Hyperparameters: grid size $N$ and the partially filled grid $\texttt{given\_grid}$ (where $0$ or \texttt{.} denotes a blank).

\paragraph{Magic Square.}
Given an $N\times N$ grid, the task is to assign integers so that every row, every column, and both main diagonals sum to the same magic constant.
Hyperparameters: order $N$, value domain (typically $\{1,\dots,N^2\}$ with an all-different constraint), and an optional set of pre-filled cells $\texttt{givens}=[(r,c,v)]$.

\paragraph{Latin Square.}
Given an $N\times N$ grid over symbols $\{1,\dots,N\}$, the task is to fill the grid so that each symbol appears exactly once in every row and every column.
Hyperparameters: order $N$ and an optional partially filled grid $\texttt{given\_grid}\in(\{0,1,\dots,N\})^{N\times N}$, where $0$ denotes blanks.

\paragraph{Futoshiki.}
Given an $N\times N$ Latin-square-style grid together with inequality clues between neighboring cells, the task is to fill the grid with symbols $\{1,\dots,N\}$ satisfying both row/column uniqueness and all inequality constraints.
Hyperparameters: order $N$, an optional partially filled grid $\texttt{given\_grid}$, and a set of directed inequality edges between adjacent cells $\texttt{ineq}=[((r_1,c_1),(r_2,c_2))]$, where each edge denotes $(r_1,c_1) < (r_2,c_2)$.

\section{Prompt Templates}
\label{app:prompt}
Below, we present the LLM template used for SAT-RL training on SAT-Bench.

\begin{tcolorbox}[
  colframe=codebg,
  colback=white,
  coltitle=blue!10!black,
  title=Input
]
\begin{lstlisting}[breaklines=true, breakatwhitespace=true, columns=fullflexible, keepspaces=true]
Solve the problem step by step and put your final answer in the format ```<answer>```.

If the evidence is genuinely insufficient after using the tool, you may output ```idk```.

For this instance, you must import and use the `{{sampled_tool}}` as the solver/tool call.

Variable Mapping:
{{variable_mapping}}

Problem Statement:
{{problem}}

\end{lstlisting}
\end{tcolorbox}

Below, we present the LLM template used for training in the other settings. This template is also used for all evaluations.

\begin{tcolorbox}[
  colframe=codebg,
  colback=white,
  coltitle=blue!10!black,
  title=Input
]
\begin{lstlisting}[breaklines=true, breakatwhitespace=true, columns=fullflexible, keepspaces=true]
Solve the problem step by step and put your final answer in the format ```<answer>```.

Problem Statement:
{{problem}}

\end{lstlisting}
\end{tcolorbox}

\section{OR-Tools Example}
\label{app:example_ortools}
Below is an example OR-Tools program.
\begin{lstlisting}[breaklines=true, breakatwhitespace=true, columns=fullflexible, keepspaces=true]
from ortools.sat.python import cp_model

class Counter(cp_model.CpSolverSolutionCallback):
    def __init__(self, x, y, want_print=True):
        super().__init__()
        self._x, self._y = x, y
        self.count = 0
        self.want_print = want_print

    def on_solution_callback(self):
        self.count += 1
        if self.want_print:
            print(f"Solution {self.count}: x={self.Value(self._x)}, y={self.Value(self._y)}")

model = cp_model.CpModel()

# Variables (nonnegative integers)
x = model.NewIntVar(0, 50, "x")   # upper bounds can be loose
y = model.NewIntVar(0, 50, "y")

# Constraint: 4x + 6y = 50
model.Add(4 * x + 6 * y == 50)

solver = cp_model.CpSolver()
cb = Counter(x, y, want_print=True)
solver.SearchForAllSolutions(model, cb)

print("Number of solutions:", cb.count)

\end{lstlisting}

\section{Solved Examples}
\label{app:solved_examples}
Below, we show examples solved during training and evaluation.

\subsection{Solved SAT-Bench Example}
\label{app:solved_satbench}
Below, we show an example SAT-Bench instance successfully solved by Qwen3.5-9B during SAT-RL training. This instance requires the use of the Z3 solver.
We omit the repeated system prompt template and directly show the instance below.

\textbf{Input.} The input instance is as follows: 
\begin{lstlisting}[basicstyle=\ttfamily\small, breaklines=true, breakatwhitespace=true, columns=fullflexible, keepspaces=true]

Scenario:
A team of three scientists—Anna, Ben, and Clara—are conducting research on two phenomena (0 = gravitational waves, 1 = dark matter). They conduct each experiment in two different environments (0 = laboratory, 1 = field). For each scientist, phenomenon, and environment, we decide whether the experiment was successfully conducted. Each decision is independent.

Conditions:
1. Either Clara successfully conducts an experiment on dark matter in the field, or Anna does not successfully conduct an experiment on dark matter in the laboratory.
2. Either Clara successfully conducts an experiment on dark matter in the field, or Ben does not successfully conduct an experiment on gravitational waves in the laboratory.
3. Either Clara does not successfully conduct an experiment on dark matter in the field, or Ben does not successfully conduct an experiment on gravitational waves in the laboratory.
4. Either Clara successfully conducts an experiment on dark matter in the field, or Ben successfully conducts an experiment on gravitational waves in the laboratory.

Question:
Can all these experimental conditions be simultaneously satisfied?

\end{lstlisting}

\textbf{Output.} The model's output is as follows: 
\begin{lstlisting}[basicstyle=\ttfamily\small, breaklines=true, breakatwhitespace=true, columns=fullflexible, keepspaces=true]
<think>
Let me analyze this problem step by step.

First, I need to understand the variables:
- x(i, j, k) means scientist i successfully conducts an experiment on phenomenon j in environment k
- i in {0, 1, 2} (Anna, Ben, Clara)
- j in {0, 1} (gravitational waves, dark matter)
- k in {0, 1} (laboratory, field)

Now let me translate the conditions:

Condition 1: Either Clara successfully conducts an experiment on dark matter in the field, or Anna does not successfully conduct an experiment on dark matter in the laboratory.
- Clara = i=2, dark matter = j=1, field = k=1: x(2, 1, 1)
- Anna = i=0, dark matter = j=1, laboratory = k=0: x(0, 1, 0)
- Condition 1: x(2, 1, 1) ∨ ¬x(0, 1, 0)

Condition 2: Either Clara successfully conducts an experiment on dark matter in the field, or Ben does not successfully conduct an experiment on gravitational waves in the laboratory.
- Clara = i=2, dark matter = j=1, field = k=1: x(2, 1, 1)
- Ben = i=1, gravitational waves = j=0, laboratory = k=0: x(1, 0, 0)
- Condition 2: x(2, 1, 1) ∨ ¬x(1, 0, 0)

Condition 3: Either Clara does not successfully conduct an experiment on dark matter in the field, or Ben does not successfully conduct an experiment on gravitational waves in the laboratory.
- Clara = i=2, dark matter = j=1, field = k=1: x(2, 1, 1)
- Ben = i=1, gravitational waves = j=0, laboratory = k=0: x(1, 0, 0)
- Condition 3: ¬x(2, 1, 1) ∨ ¬x(1, 0, 0)

Condition 4: Either Clara successfully conducts an experiment on dark matter in the field, or Ben successfully conducts an experiment on gravitational waves in the laboratory.
- Clara = i=2, dark matter = j=1, field = k=1:  x(2, 1, 1)
- Ben = i=1, gravitational waves = j=0, laboratory = k=0: x(1, 0, 0)
- Condition 4: x(2, 1, 1) ∨ x(1, 0, 0)

Let me use Z3 to check if these conditions can be simultaneously satisfied.

For Z3, I'll use x_2_1_1 for x(2,1,1) and x_1_0_0 for x(1,0,0).
</think>
<tool_call>
from z3 import Bool, Or, Not, And, Solver, sat, unsat

# Define the variables
x_2_1_1 = Bool("x_2_1_1")  # Clara, dark matter, field
x_1_0_0 = Bool("x_1_0_0")  # Ben, gravitational waves, laboratory

# Create the conditions
condition1 = Or(x_2_1_1, Not(x_0_1_0)) if False else Or(x_2_1_1, Not(Bool("x_0_1_0")))
condition2 = Or(x_2_1_1, Not(x_1_0_0))
condition3 = Or(Not(x_2_1_1), Not(x_1_0_0))
condition4 = Or(x_2_1_1, x_1_0_0)

# Create solver and add constraints
solver = Solver()
solver.add(And(condition2, condition3, condition4))

# Check satisfiability
result = solver.check()
print(result)
</tool_call>
<tool_response>
stdout: sat
</tool_response>
<think>
The solver returned "sat" which means the conditions are satisfiable. This means all conditions can be simultaneously satisfied.

Therefore, the answer is yes.
</think>

```yes```
\end{lstlisting}
\textbf{Reward.} The reward score is as follows:
\begin{lstlisting}[breaklines=true, breakatwhitespace=true, columns=fullflexible, keepspaces=true]
{"format": 0.1, "correctness": 1.0, "tool_success": 0.1, "equivalence": 1.0}
\end{lstlisting}
\textbf{Justification.}
The model's response contains a correctly parsed answer and a compilable Z3 tool call.
The variable mapping is as follows: $x_{2,1,1}$ is mapped to ``Clara conducts the dark-matter experiment in the field'', $x_{1,0,0}$ is mapped to ``Ben conducts the gravitational-wave experiment in the laboratory'', and $x_{0,1,0}$ is mapped to ``Anna conducts the dark-matter experiment in the laboratory''.

The ground-truth formula is $F_{\text{truth}}=\left(x_{2,1,1} \lor \neg x_{0,1,0}\right)
\land
\left(x_{2,1,1} \lor \neg x_{1,0,0}\right)
\land
\left(\neg x_{2,1,1} \lor \neg x_{1,0,0}\right)
\land
\left(x_{2,1,1} \lor x_{1,0,0}\right)$.

The extracted formula is 
\(F_{\text{pred}}=
\left(x_{2,1,1} \lor \neg x_{1,0,0}\right)
\land
\left(\neg x_{2,1,1} \lor \neg x_{1,0,0}\right)
\land
\left(x_{2,1,1} \lor x_{1,0,0}\right)
\).
Although $F_{\text{pred}}$ does not include the clause $\left(x_{2,1,1} \lor \neg x_{0,1,0}\right)$, it is still logically equivalent to $F_{\text{truth}}$ (that is, $F_{\text{truth}}\leftrightarrow F_{\text{pred}}$). In fact, the truth table is \[
\begin{array}{c c c|c c}
x_{2,1,1} & x_{1,0,0} & x_{0,1,0} & F_1 & F_2 \\
\hline
0 & 0 & 0 & 0 & 0 \\
0 & 0 & 1 & 0 & 0 \\
0 & 1 & 0 & 0 & 0 \\
0 & 1 & 1 & 0 & 0 \\
1 & 0 & 0 & 1 & 1 \\
1 & 0 & 1 & 1 & 1 \\
1 & 1 & 0 & 0 & 0 \\
1 & 1 & 1 & 0 & 0
\end{array}
\]

\subsection{Solved Enigmata Example}
\label{app:solved_enigmata}
\textbf{Input.} The input instance is as follows: 
\begin{lstlisting}[breaklines=true, breakatwhitespace=true, columns=fullflexible, keepspaces=true]


1. **NOTATIONS:**
    The initial board consists of empty cells and blocked cells.
    Empty cells are denoted by '.', blocked cells are denoted by 'X', and stars are denoted by '*'.
2. **STAR MUST BE PLACED IN EMPTY CELL:**
    Each star must be placed in an EMPTY cell.
    Blocked cells cannot contain stars.
    You can only change cells denoted by '.', and must not change cells denoted by 'X'.
3. **EXACTLY 1 STAR IN EACH ROW AND COLUMN:**
    Each row and column must contain EXACTLY one star.
    No two stars can be in the same row or column.
    There shouldn't be rows or columns without stars.
4. **NO ADJACENT STARS ROW-WISE, COLUMN-WISE, OR DIAGONALLY:**
    No two stars can be adjacent to each other, even diagonally.
    - Row-wise adjacency: two stars are in the same row, and there is no empty cell between them.
    - Column-wise adjacency: two stars are in the same column, and there is no empty cell between them.
    - Diagonal adjacency: two stars are in the same diagonal, and there is no empty cell between them.
    Note that you are prone to make mistakes in diagonal adjacency, so be extra careful. Here are 2 examples of diagonal adjacency in a partial grid:
    ```
    . * . .
    . . * .
    ```
    ```
    . . * .
    . * . .
    ```
    In both cases, the two stars are diagonally adjacent. You should avoid this situation.
5. **CHECK FOR CONSTRAINTS AND BACKTRACK:** 
    In each step, you should check if it violates the constraints in 2., 3., and 4.
    If you find inconsistencies, you should backtrack and try a different placement.
    If you find that you can't place a star anywhere without violating the constraints, you should backtrack and try a different placement.
    Since there is only one solution, there is great chance your first placement is incorrect, which means you might need to start over.
6. **RESPONSE FORMAT:** 
    Your response should include the final board.
    The final board is a revised version of the initial board in a way that if you need to place a star in an empty cell, replace the '.' with a '*'.
    Don't change the blocked cells denoted by 'X'.
    Your final board should be wrapped between <begin_board> and <end_board> tags.
    Use the following structure:

    Final Board:
    ```
    <begin_board>
    [Final Board]
    <end_board>
    ```

Here is the puzzle:
. X X . X . .
. X X . . X .
. X X . X X X
. . . . . . .
. . . . X X .
. . X . X X X
. . . . . . .
\end{lstlisting}
\textbf{Output.} The model's output is as follows: 
\begin{lstlisting}[basicstyle=\ttfamily\small, breaklines=true, breakatwhitespace=true, columns=fullflexible, keepspaces=true]
<think>
The user wants me to solve a star-battle puzzle.

Let's start by setting up the model with pysat, then solve, then print the result:

Step 1: Import pysat and define binary variables `star[r][c]`.
Step 2: Add constraints for each row/col to have exactly one star (clauses like `sum(star[r][c] for all c) == 1`).
Step 3: Add constraints to prevent adjacent/diagonal stars (no variable equals any neighbor in the 3x3 area).
Step 4: Add constraints on blocked cells (each star variable for a blocked cell must be false).
Step 5: Solve the instance.
Step 6: Extract the positions where `star[r][c]` is true and construct the final board.
</think>

<tool_call>
from pysat.formula import CNF, IDPool
from pysat.solvers import Glucose3


def solve():
    puzzle = [
        [".", "X", "X", ".", "X", ".", "."],
        [".", "X", "X", ".", ".", "X", "."],
        [".", "X", "X", ".", "X", "X", "X"],
        [".", ".", ".", ".", ".", ".", "."],
        [".", ".", ".", ".", "X", "X", "."],
        [".", ".", "X", ".", "X", "X", "X"],
        [".", ".", ".", ".", ".", ".", "."],
    ]
    rows = len(puzzle)
    cols = len(puzzle[0])
    vpool = IDPool()
    stars = [[vpool.id(f"star_{r}_{c}") for c in range(cols)] for r in range(rows)]
    cnf = CNF()
    for r in range(rows):
        cnf.append(list(stars[r]))
        for c in range(cols):
            cnf.append([-stars[r][c]])
    for c in range(cols):
        cnf.append(list(stars[r][c] for r in range(rows)))
    for r in range(rows):
        for c in range(cols):
            for dr in (-1, 0, 1):
                for dc in (-1, 0, 1):
                    if dr == 0 and dc == 0:
                        continue
                    nr, nc = r + dr, c + dc
                    if 0 <= nr < rows and 0 <= nc < cols:
                        cnf.append([-stars[r][c], -stars[nr][nc]])
    for r in range(rows):
        for c in range(cols):
            if puzzle[r][c] == "X":
                cnf.append([-stars[r][c]])

    with Glucose3(bootstrap_with=cnf.clauses) as solver:
        if not solver.solve():
            return "Unsolvable"
        model = set(solver.get_model())

    planted = []
    for r in range(rows):
        for c in range(cols):
            if stars[r][c] in model:
                planted.append((r, c))
    final_board = []
    for r in range(rows):
        row_chars = []
        for c in range(cols):
            if puzzle[r][c] == "X":
                row_chars.append("X")
            elif (r, c) in planted:
                row_chars.append("*")
            else:
                row_chars.append(".")
        final_board.append("".join(row_chars))
    return "\n".join(final_board)


print(solve())
</tool_call>
<tool_response>
stdout: Unsolvable
</tool_response>
<think>
The previous solver returned "Unsolvable", which should not be possible for a well-formed puzzle with a unique solution. I must have made an error in how the constraints were modeled—specifically the adjacency constraint may be too restrictive or incorrectly implemented. Let me carefully reconsider the diagonal adjacency model and simplify the exact-1 constraints using linear expressions, then retry the solver call.
</think>

<tool_call>
from pysat.formula import CNF, IDPool
from pysat.solvers import Glucose3
from pysat.card import CardEnc, EncType


def solve():
    puzzle = [
        [".", "X", "X", ".", "X", ".", "."],
        [".", "X", "X", ".", ".", "X", "."],
        [".", "X", "X", ".", "X", "X", "X"],
        [".", ".", ".", ".", ".", ".", "."],
        [".", ".", ".", ".", "X", "X", "."],
        [".", ".", "X", ".", "X", "X", "X"],
        [".", ".", ".", ".", ".", ".", "."],
    ]
    rows = len(puzzle)
    cols = len(puzzle[0])
    vpool = IDPool()
    stars = [[vpool.id(f"star_{r}_{c}") for c in range(cols)] for r in range(rows)]
    cnf = CNF()
    for r in range(rows):
        cnf.extend(CardEnc.equals(lits=stars[r], bound=1, vpool=vpool, encoding=EncType.seqcounter).clauses)
    for c in range(cols):
        cnf.extend(CardEnc.equals(lits=[stars[r][c] for r in range(rows)], bound=1, vpool=vpool, encoding=EncType.seqcounter).clauses)
    for r in range(rows):
        for c in range(cols):
            for dr in (-1, 0, 1):
                for dc in (-1, 0, 1):
                    if dr == 0 and dc == 0:
                        continue
                    nr, nc = r + dr, c + dc
                    if 0 <= nr < rows and 0 <= nc < cols:
                        cnf.append([-stars[r][c], -stars[nr][nc]])
    for r in range(rows):
        for c in range(cols):
            if puzzle[r][c] == "X":
                cnf.append([-stars[r][c]])

    with Glucose3(bootstrap_with=cnf.clauses) as solver:
        if not solver.solve():
            return "Unsolvable"
        model = set(solver.get_model())

    planted = []
    for r in range(rows):
        for c in range(cols):
            if stars[r][c] in model:
                planted.append((r, c))
    final_board = []
    for r in range(rows):
        row_chars = []
        for c in range(cols):
            if puzzle[r][c] == "X":
                row_chars.append("X")
            elif (r, c) in planted:
                row_chars.append("*")
            else:
                row_chars.append(".")
        final_board.append("".join(row_chars))
    return "\n".join(final_board)


print(solve())
</tool_call>
<tool_response>
stdout: .XX.X*.
.XX*.X.
*XX.XXX
..*....
....XX*
.*X.XXX
....*..
</tool_response>
<think>
The solver returned a valid solution:
```
.XX.X*.
.XX*.X.
*XX.XXX
..*....
....XX*
.*X.XXX
....*..
```

This should be formatted according to the required output structure with the <begin_board> and <end_board> tags. Let me verify this solution is correct:
- Row 0: star at column 5
- Row 1: star at column 3
- Row 2: star at column 0
- Row 3: star at column 2
- Row 4: star at column 6
- Row 5: star at column 1
- Row 6: star at column 4

This gives all columns [5,3,0,2,6,1,4], which are all distinct (0-6 exactly once). No row or column has more than one star. Checking diagonally adjacent stars:
- Row 0 col 5 and Row 1 col 3: not adjacent
- Row 1 col 3 and Row 2 col 0: not adjacent
- Row 2 col 0 and Row 3 col 2: not adjacent
- Row 3 col 2 and Row 4 col 6: not adjacent
- Row 4 col 6 and Row 5 col 1: not adjacent
- Row 5 col 1 and Row 6 col 4: not adjacent

None are adjacent horizontally, vertically, or diagonally. The placement also respects the blocked cells ('X'). This is the correct and unique solution.
</think>
Final Board:
```
<begin_board>
.XX.X*.
.XX*.X.
*XX.XXX
..*....
....XX*
.*X.XXX
....*..
<end_board>
```

\end{lstlisting}

\textbf{Justification:}
The first PySAT call creates variables for all 49 cells and adds an at-least-one clause for each row, but this formulation produces an UNSAT result that contradicts the task's stated unique-solution premise. The model then corrects the formula and obtains a satisfying solution. The star columns by row are $(5, 3, 0, 2, 6, 1, 4)$. These placements ensure that every row and column has exactly one star, that each star replaces an empty cell, and that the star columns in adjacent rows differ by more than one. Therefore, the solution is correct, as confirmed by the verifier.

\section{Additional Training Trajectories}
\begin{figure*}[ht!]
\centering
\subfigure[Total reward $r$.]{%
    \includegraphics[width=0.315\textwidth]{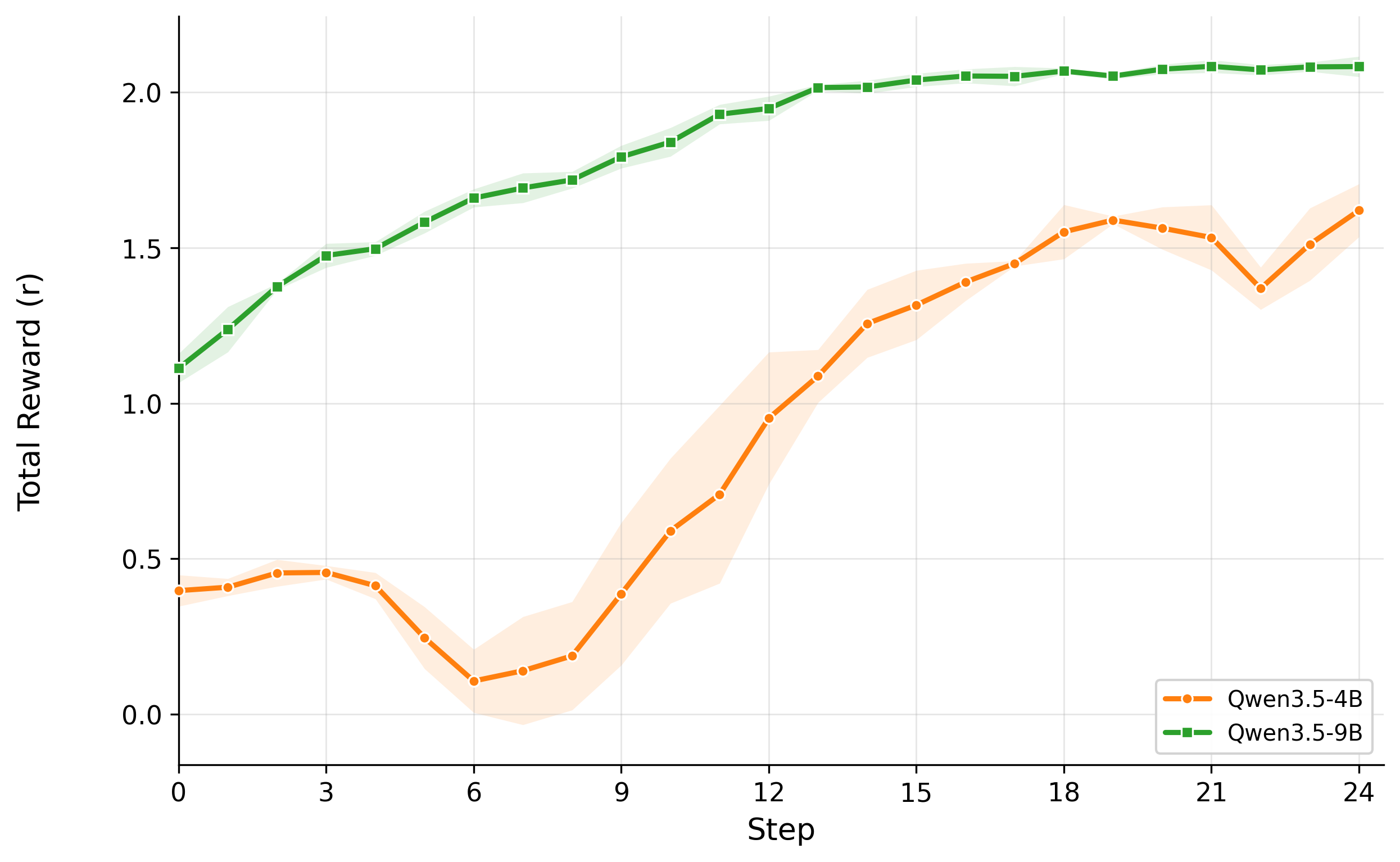}%
}\hfill
\subfigure[Validation correctness.]{%
    \includegraphics[width=0.315\textwidth]{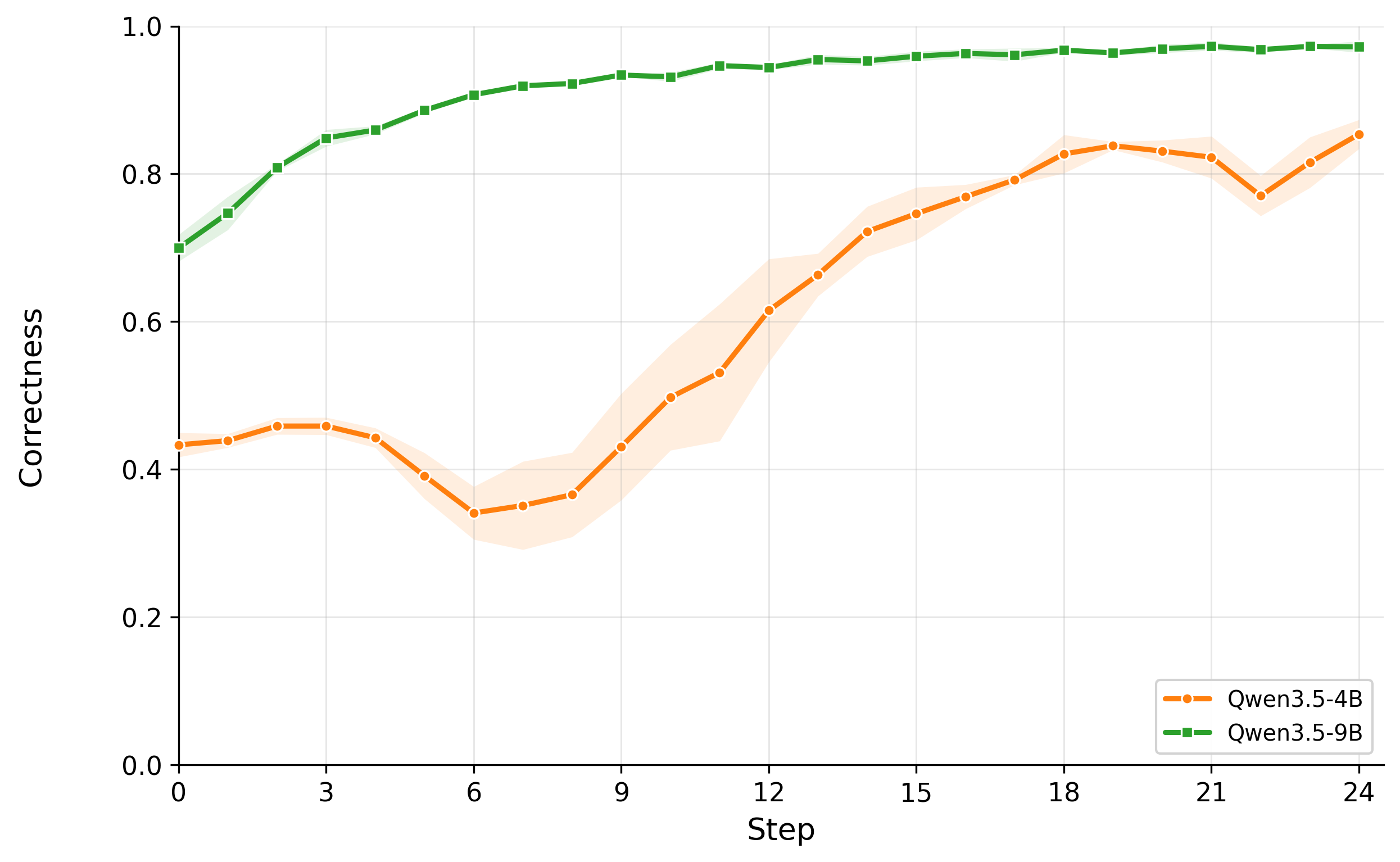}%
}\hfill
\subfigure[Policy entropy.]{%
    \includegraphics[width=0.315\textwidth]{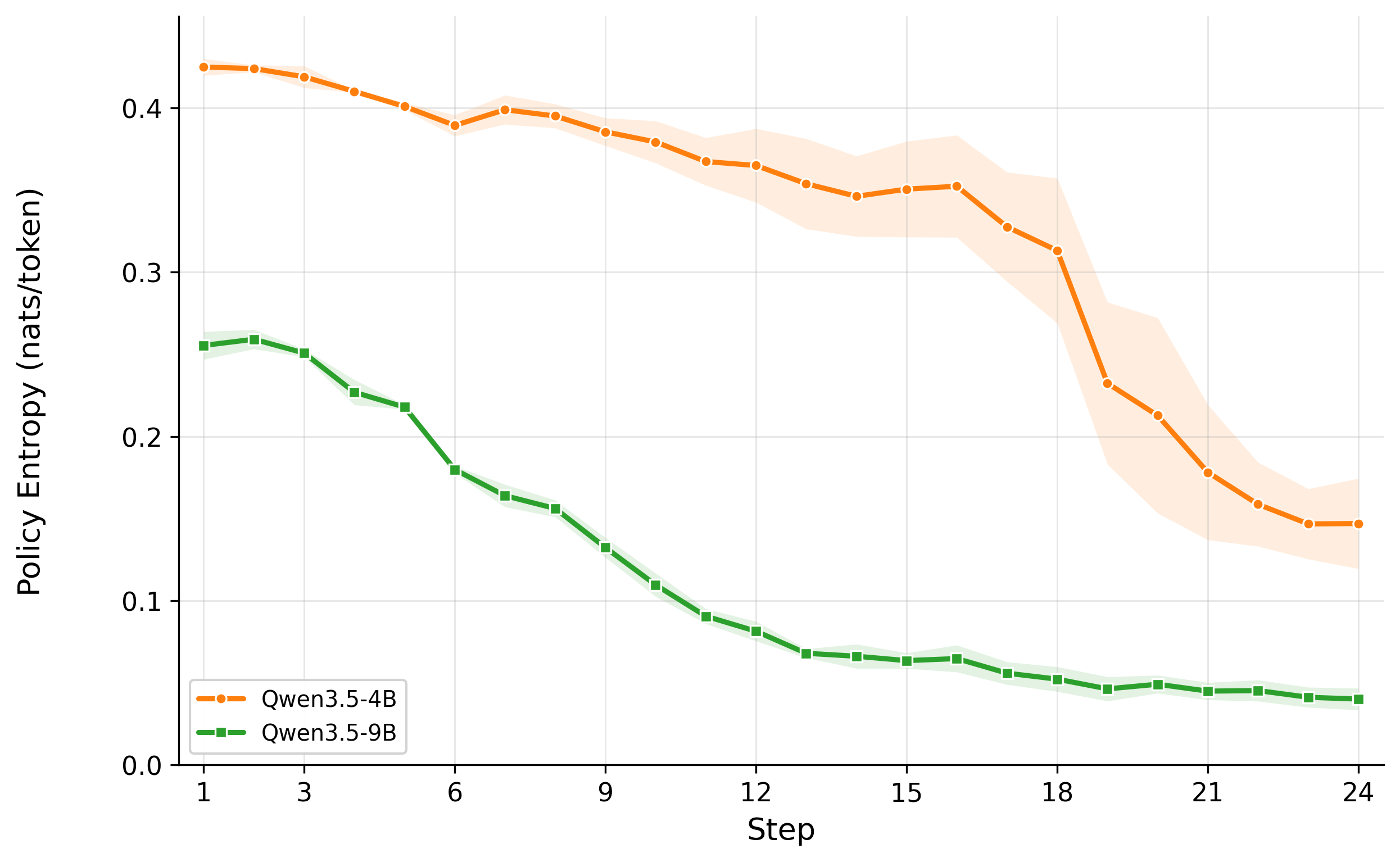}%
}

\par\medskip
\subfigure[Adjusted correctness reward $r_{\text{correct}}$.]{%
    \includegraphics[width=0.43\textwidth]{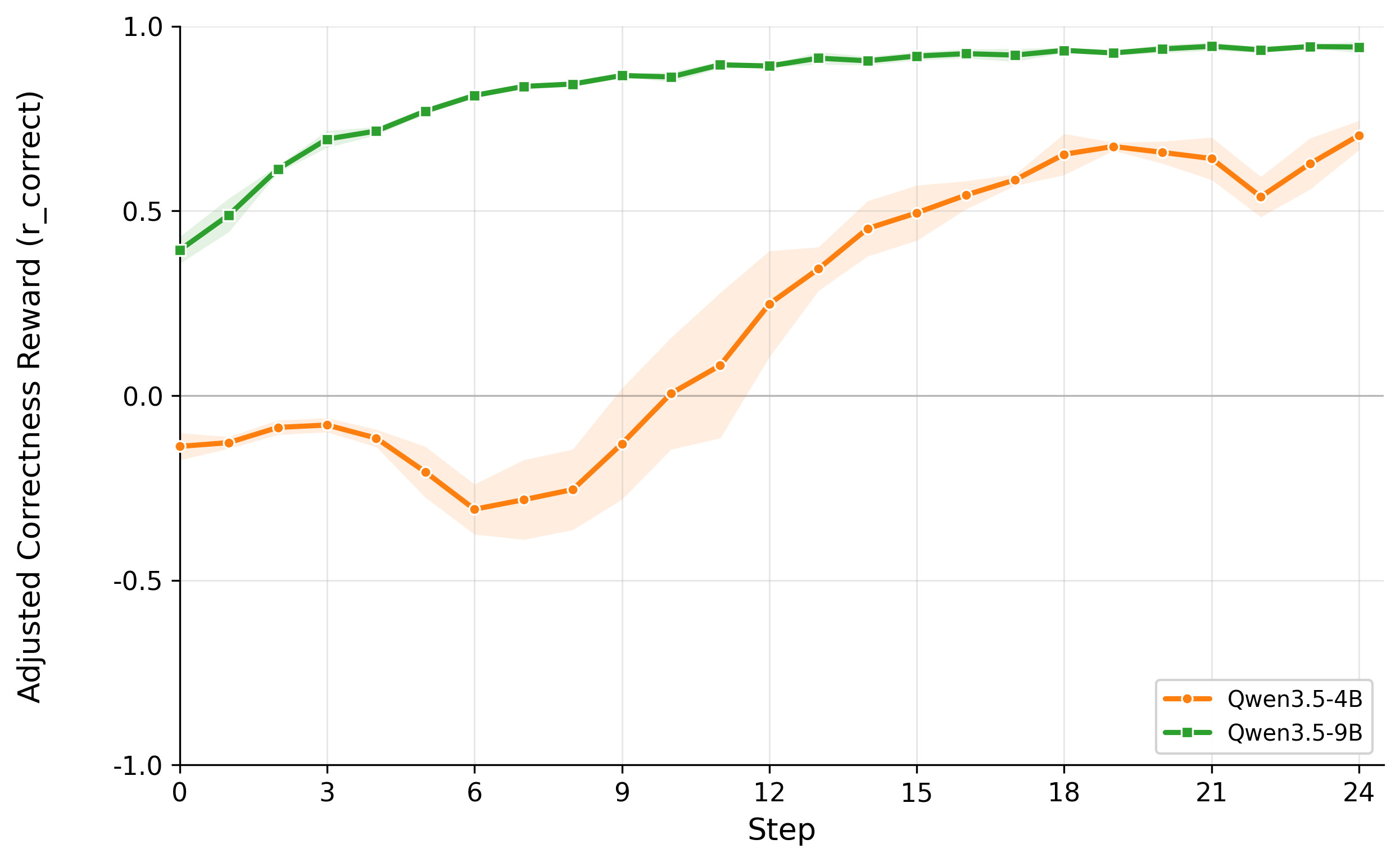}%
}\hspace{0.03\textwidth}
\subfigure[Format reward $r_{\text{format}}$.]{%
    \includegraphics[width=0.43\textwidth]{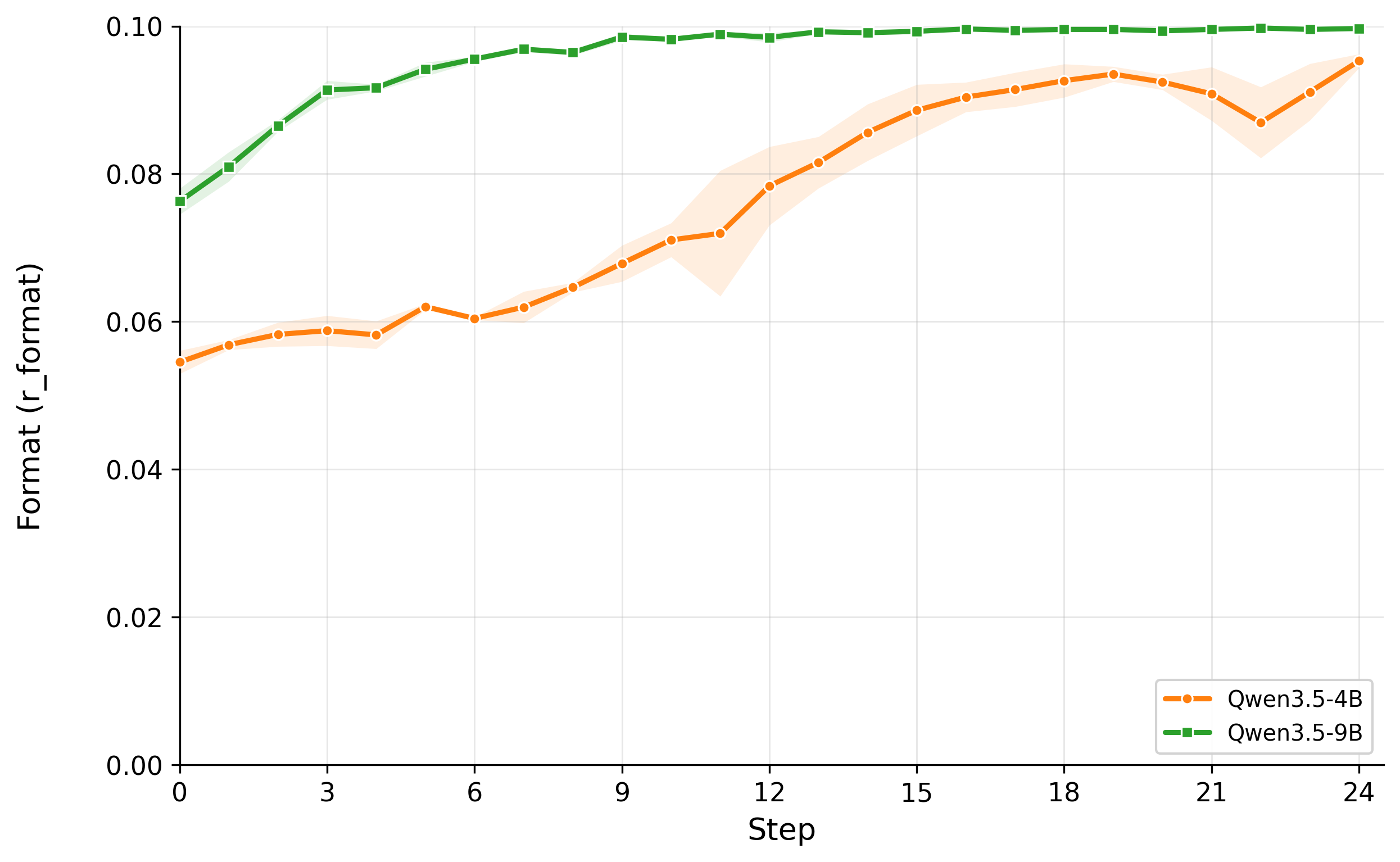}%
}

\par\medskip
\subfigure[Tool reward $r_{\text{tool}}$.]{%
    \includegraphics[width=0.43\textwidth]{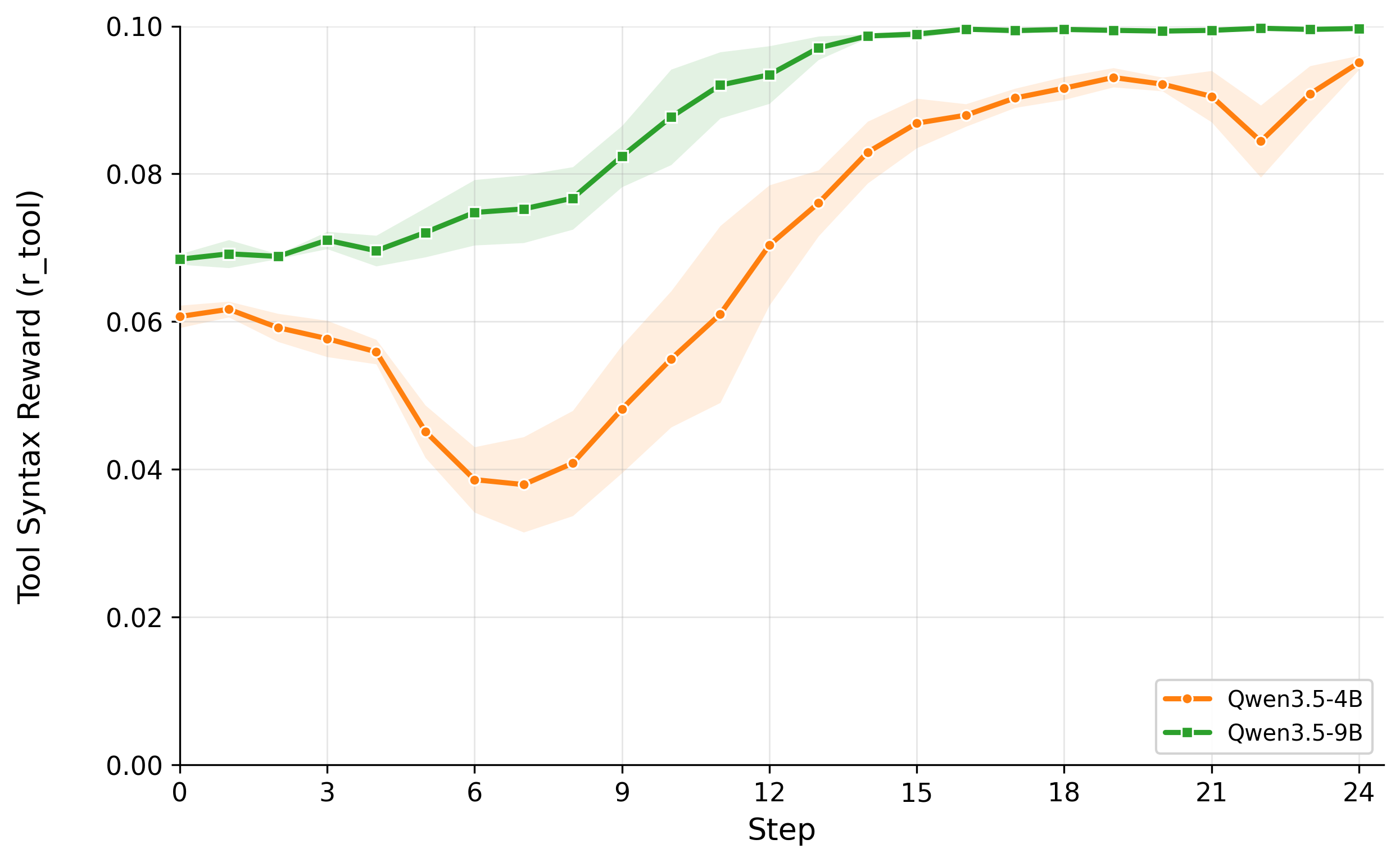}%
}\hspace{0.03\textwidth}
\subfigure[Logic-equivalence reward $r_{\text{equiv}}$.]{%
    \includegraphics[width=0.43\textwidth]{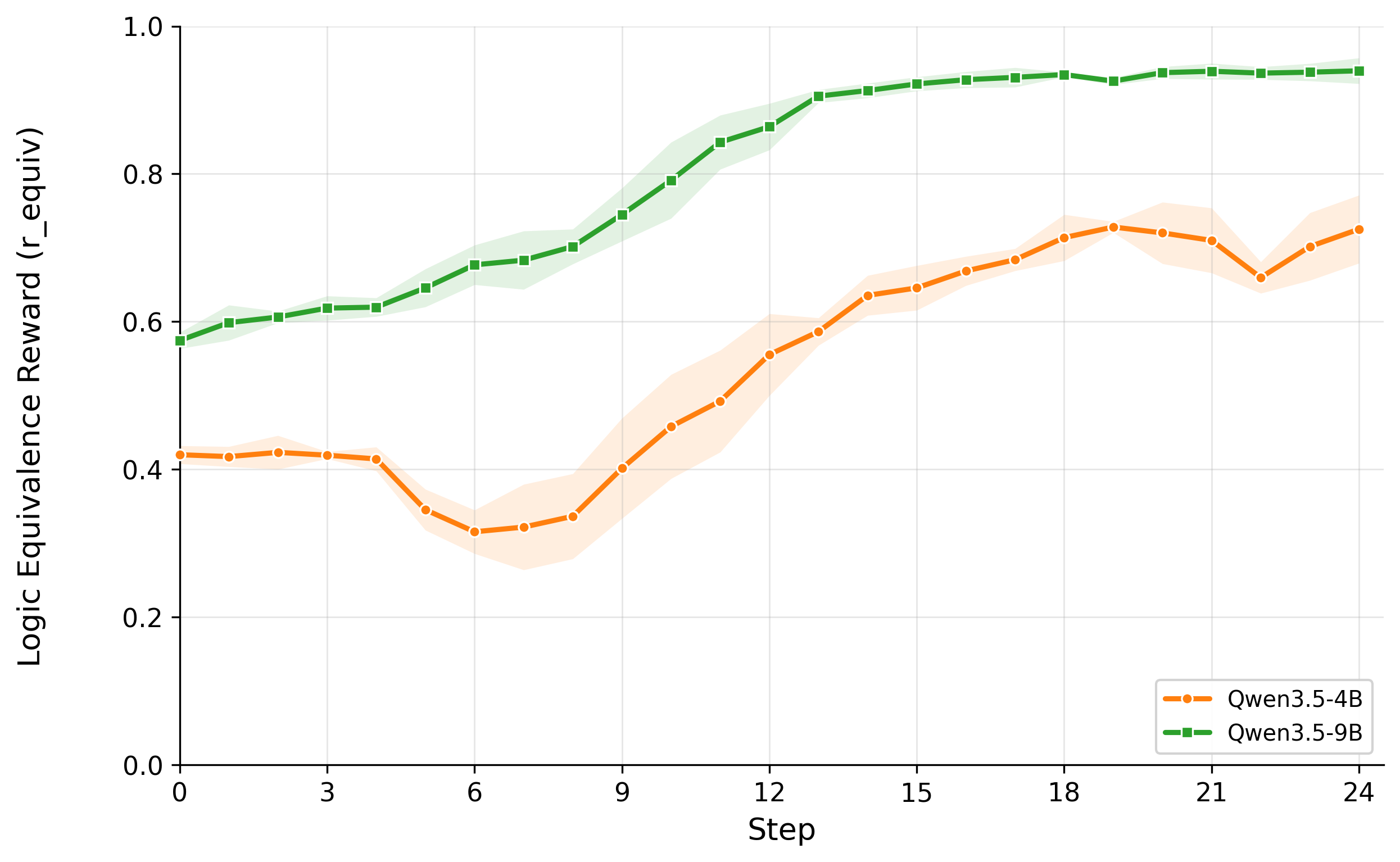}%
}

\caption{SAT-RL training trajectories on SAT-Bench for Qwen3.5-4B and Qwen3.5-9B. Curves show means over three runs; shading shows population standard deviation.}
\label{fig:appendix_training_trajectories}
\end{figure*}

Additional training trajectories of detailed reward component, the test correctness and entropy is shown to Figure~\ref{fig:appendix_training_trajectories}.

\end{document}